\documentclass{article}

\usepackage{microtype}
\usepackage{graphicx}
\usepackage{subfigure}
\usepackage{booktabs} % for professional tables
\usepackage{multirow}
\usepackage{bbding}
\usepackage{bibentry}
\usepackage{hyperref}
\usepackage{amsmath,amsfonts,amsthm,bm}

\usepackage[accepted]{icml2021}

\newtheorem{theorem}{Theorem}

\icmltitlerunning{\textsc{STONet}: A Neural-Operator-Driven Spatio-temporal Network}

\begin{document}

\twocolumn[
\icmltitle{STONet: A Neural-Operator-Driven Spatio-temporal Network}

% It is OKAY to include author information, even for blind
% submissions: the style file will automatically remove it for you
% unless you've provided the [accepted] option to the icml2021
% package.

% List of affiliations: The first argument should be a (short)
% identifier you will use later to specify author affiliations
% Academic affiliations should list Department, University, City, Region, Country
% Industry affiliations should list Company, City, Region, Country

% You can specify symbols, otherwise they are numbered in order.
% Ideally, you should not use this facility. Affiliations will be numbered
% in order of appearance and this is the preferred way.
\icmlsetsymbol{equal}{*}

\begin{icmlauthorlist}
\icmlauthor{Haitao Lin,}{westlake}
\icmlauthor{Guojiang Zhao,}{westlake}
\icmlauthor{Lirong Wu,}{westlake}
\icmlauthor{Stan Z. Li}{westlake}
\end{icmlauthorlist}

\icmlaffiliation{westlake}{Center for Artificial Intelligence Research and Innovation}

\icmlcorrespondingauthor{Haitao Lin}{linhaitao@westlake.edu.cn}
\icmlcorrespondingauthor{Stan Z. Li}{stanzqli@westlake.edu.cn}

% You may provide any keywords that you
% find helpful for describing your paper; these are used to populate
% the "keywords" metadata in the PDF but will not be shown in the document
\icmlkeywords{Machine Learning, ICML}

\vskip 0.3in
]

\begin{abstract}
Graph-based spatio-temporal neural networks are effective to model the spatial dependency among discrete points sampled irregularly from unstructured grids, thanks to the great expressiveness of graph neural networks.
However, these models are usually spatially-transductive -- only fitting the signals for discrete spatial nodes fed in models but unable to generalize to `unseen' spatial points with zero-shot. 
In comparison, for forecasting tasks on continuous space such as temperature prediction on the earth's surface, the \textit{spatially-inductive} property allows the model to generalize to any point in the spatial domain, demonstrating models' ability to learn the underlying mechanisms or physics laws of the systems, rather than simply fit the signals.
Besides, in temporal domains, \textit{irregularly-sampled} time series, e.g. data with missing values, urge models to be temporally-continuous. 
Motivated by the two issues, we propose a spatio-temporal framework based on neural operators for PDEs, which learn the underlying mechanisms governing the dynamics of spatially-continuous physical quantities.
Experiments show our model's improved performance on forecasting spatially-continuous physic quantities, and its superior generalization to unseen spatial points and ability to handle temporally-irregular data.
\end{abstract}

\vspace*{-2em}
\section{Introduction}

Studying the spatio-temporal patterns of physical quantities is of great scientific interest.
Significant progress has been achieved thanks to immense research efforts in deep neural networks for modeling the spatial dependency and temporal dynamics \citep{shi2015convolutional,guo2019attention,Zhao2020TGCN,bai2020adaptive,li2021dynamic}. 
Most of them are established for spatially-discrete nodes, such as sensors' signals of traffic flow located on discretized roads, and graph neural networks (GNNs) are usually employed to handle signals with spatially-irregular distribution and establish dependency between nodes \citep{seo2016gcgru,yu2018STGCN,li2018diffusion,rozemberczki2021pytorch}. 
\begin{figure}[]
    \centering
    \includegraphics[width=1.02\linewidth]{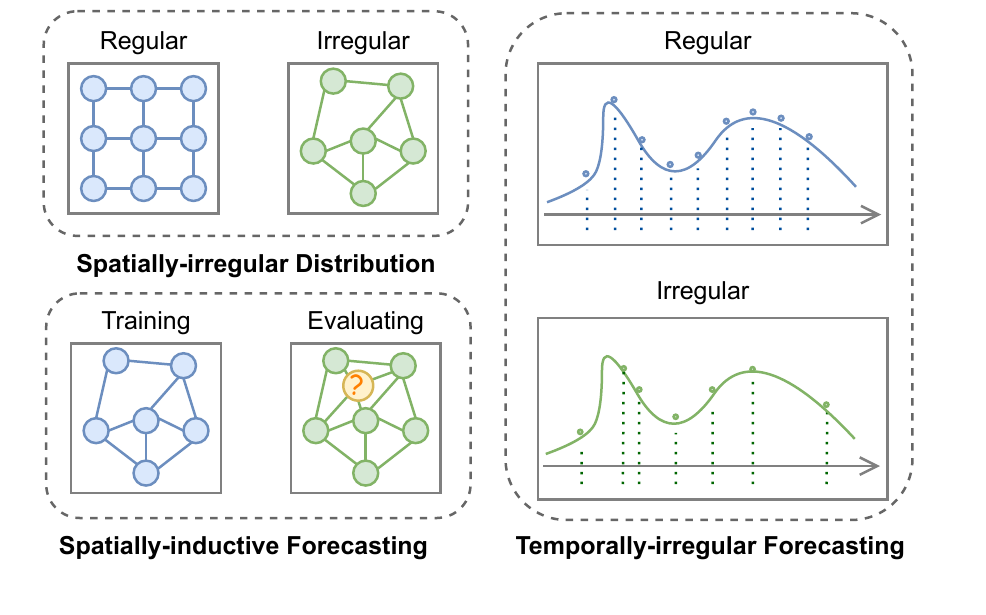}
    \vspace{-3em}
    \caption{The discussed challenges in spatio-temporal forecasting.}
    \label{fig:problemdef}
    \vspace{-2em}
\end{figure}
In comparison, models established for spatio-temporal forecasting tasks for spatially-continuous physical quantities are still rare, although in real-world scenarios, the needs for them are urgent.
Examples arise in fields like geophysics such as temperature and humidity forecasting \cite{rasp2020weatherbench}, where there exists a value of temperature or humidity at every point on the earth's surface, and acoustic or electromagnetism such as wave equation modeling \cite{saha2021physicsincorporated}.
While the previous models for discrete spatial domain can also be used for tasks like weather forecasting \cite{lin2021conditional}, they are limited to only capturing dynamics of the discrete sample points which are fed in models. However, for points in the continuous spatial domain which have not been seen by the model, they fail to generalize on them well. 
Besides, in real-world application, time intervals between observations may be non-unifrom, such as missing data scenarios, which drives us to construct a model to deal with temporal irregularly-sampled data.
% most of them only support generating predictions of uniform time interval, driving us to construct a model to deal with temporal irregularly-sampled data.

In summary, we conclude three challenges for continuous spatio-temporal models as shown in Fig.~\ref{fig:problemdef}:  (1) modeling the dependency among sample points of \textit{spatially-irregular distribution}, which can be well-solved by GNNs; (2) \textit{spatially-inductive forecasting} and (3) \textit{temporally-irregular forecasting} which are both not been well resolved.
To address (2), we aim to introduce a forecasting model that can both predict physical quantities or signals on irregularly-distributed points, and accurately generalize the learned dynamic patterns to the `unseen' points sampled from the continuous spatial domain.
Besides, when the time intervals between observations are not equal as (3) discusses, e.g. there are missing timestamps in training data, we want the established model to be continuous in temporal domains and thus can still generate accurate predictions.

Recently, great progress has been achieved in solving partial differential equations (PDEs) \citep{RAISSI2019686,jiang2020meshfreeflownet,greenfeld2019learning,kochkov2021machine}, which is able to learn the implicit or explicit mechanisms governing the dynamics of spatially-continuous physical quantities.
Inspired by this, we deduce that such models can be well generalized to the unseen spatial points and irregularly-sampled timestamps based on the fact that the dynamics governed by PDE models are applicable to any spatial point and time in the domains, and thus build a spatio-temporal model to learn the underlying PDEs for solving issue (2) and (3). 
On contrary to approaches designed to model one specific instance of PDE \citep{e2017deep,bar2019unsupervised,smith2020eikonet,shaowu2020,Raissi2020sceience}, our model aims to not be limited to one physical process and approximate different dynamical systems well. 
Therefore, neural operators \citep{lulu2021deeponet,bhattacharya2021model,li2020multipole,li2021fourier, nelsen2021random} is a desirable option, which directly learn the mapping between infinite-dimensional spaces of functions, requiring no knowledge of underlying PDEs, and only data.
We name our model as spatio-temporal operator net (\textsc{STONet}), which first encodes the history observations with graph neural operators, and generates future predictions with universal operators. The multipole-graph-based encoder allows the spatial points to be \textit{irregularly distributed} on unstructured grids, and the neural-operator-driven architecture enables both \textit{spatially-inductive} and \textit{temporally-irregular forecasting}. Our primary contributions include:
\vspace{-0.3cm}
\begin{itemize}
    \item We set up an encoder with historical observations fed in, based on multipole graph operators, which can capture the long-range spatial dependency as well as approximate the mappings of learned parametric functions into solution representation functions. (Sec. \ref{sec:graphkerenc})
    \item We establish a universal-operator-based decoder, approximating solution representation functions with historical observations, for future predictions. (Sec. \ref{sec:deepoperdec})
    \item We show our model's high performance for spatio-temporal forecasting on real-world datasets, points in which are irregularly distributed in continuous spatial domains. Besides, experimental results prove its capability of spatially-inductive and temporally-irregular forecasting with further analysis. (Sec. \ref{sec:experiment})
\end{itemize}

\section{Background}
\subsection{Notation and Preliminary}
Let $\mathrm{F}$ be a multivariate function, and the partial differential equation governing the continuous dynamical system reads
\begin{align}
    \label{eq:diffeq}
    \frac{\partial u}{\partial t} = 
    \mathrm{F}\left (\bm{\mathrm{x}},  \frac{\partial u}{\partial x_1},\ldots,\frac{\partial u}{\partial x_m}, \frac{\partial^2 u}{\partial x_1^2}, \frac{\partial^2 u}{\partial x_1\partial x_2}, \frac{\partial^2 u}{\partial x_2^2}, \ldots; \theta\right ),
\end{align}
where  $u = u(\bm{\mathrm{x}}, t) \in \mathcal{U}$ is the observed physical quantity of $m$-dimensional spatial location $\bm{\mathrm{x}} = (x_1, \ldots, x_m) \in D$ at time $t\in[0, T]$ with $\mathcal{U}$  a separable Banach space., and $\theta = \theta(\bm{\mathrm{x}}, t)$ reflects the external influence. 
For example, in homogeneous heat equation,
$   \frac{\partial u}{\partial t} = \sum_{i=1}^{m} \frac{\partial^2 u}{\partial x_i^2} + q$
where $u$ is the temperature, and $\theta = q$ is the heat sources.
Note that when the equation describes the dynamics of time, we write $u(\bm{\mathrm{x}}, t)$ as $u_{\bm{\mathrm{x}}} (t)$ for fixed spatial location, and as $u_{t} (\bm{\mathrm{x}})$ when the equation or operator is with respect to $\bm{\mathrm{x}}$. 
For example, the heat equation can be written as
$\frac{\partial u_{\mathrm{x}}}{\partial t} = \sum_{i=1}^{m} \frac{\partial^2 u_t}{\partial x_i^2} + q$.

\subsection{Neural Operator for PDE}
\label{sec:neuraloperator}
\paragraph{Parametric PDEs.} Assume that the term on the right side of the Eq.~\ref{eq:diffeq} can be parameterized as a parametric PDE, and there exists $f(\bm{\mathrm{x}},t)$, such that 
\vspace{-0.2cm}
\begin{align}\label{eq:spacediffeq}
    (\mathcal{\bm{L}}_{a_t} u_t)(\bm{\mathrm{x}}) = f(\bm{\mathrm{x}},t),
\end{align}
where $\mathcal{\bm{L}}_{a_t}$ is a differential operator with respect to $\bm{\mathrm{x}}$, determined by $a_t(\bm{\mathrm{x}}) \in \mathcal{A}$ with $\mathcal{A}$ a separable Banach space.
Because the operator is not with respect to $t$, $t$ can be regarded as a parameter in $u$ and $a$, rather than a variable like $\bm{\mathrm{x}}$. When the parametric function $a_t$ differs as time varies, $u_t(\bm{\mathrm{x}})$ also changes over time.

To obtain the solution $u_t(\bm{\mathrm{x}})$, we aim to approximate a target mapping between two infinite-dimensional function spaces, which is $\mathcal{F}: \mathcal{A} \rightarrow \mathcal{U}$, such that $\mathcal{F}(a_t) = u_t$. Given the $n_t \times n_s$ observations $\{\left(a_{t_{i}}(\bm{\mathrm{x}}_{j}), u_{t_{i}}(\bm{\mathrm{x}}_{j})\right): i=1,\ldots,n_t, j=1,\ldots,n_s\}$, $\mathcal{F^\dag}$ is to used to approximate $\mathcal{F}$ such that $\mathcal{F^\dag}\approx \mathcal{F}$. 
\vspace{-0.25cm}
\paragraph{Universal operators.} When the parametric function $a(\bm{\mathrm{x}},t)$  or equation formulation is totally unknown, a universal approximator can be used to directly learn the operator. For a certain $u(\bm{\mathrm{x}},t)$, the target operator is $\mathcal{G}$, given $n_t \times n_s$ observations $\{\left(u(\bm{\mathrm{x}}_{j}, t_{i}), \mathcal{G}(u)(\bm{\mathrm{x}}_{j}, t_{i})\right): i=1,\ldots,n_t, j=1,\ldots,n_s\}$, we try to establish a universal approximator $\mathcal{G^\dag}$, such that
$\mathcal{G^\dag}\approx\mathcal{G}$, according to the provided observations. As such, we can obtain $\mathcal{G^\dag}(u)(\bm{\mathrm{x}},t)$ at any $\bm{\mathrm{x}} \in D$ and $t \in [0, T]$ as its approximation. 

Here are the differences between $\mathcal{F}$ and $\mathcal{G}$. $\mathcal{F}$ learns a solution mapping shared by PDEs parameterized by a family of operators $\{\mathcal{L}_{a_t}\}_{t\in [0,T]}$ with the same parametric forms $\mathcal{L}$ and different parameters $\{a_t\}$. For example, Eq.~\ref{eq:spacediffeq} governed by the second order elliptic operator $\mathcal{L}_a\cdot = -\mathrm{div}(a\nabla\cdot)$ describes many physical phenomenons including hydrology \cite{TransportPhenomena} and elasticity\cite{Elasticity}.  In comparison, the second one approximates a single operator $\mathcal{G^\dag}$, which maps a certain function $u$ to $\mathcal{G^\dag}(u)$.

\subsection{Spatio-temporal Forecasting}
\label{sec:spatemp}
%  Given $n_s$ physical quantities located on the plane or sphere at time  $t$. 
Given $n_s$ fixed spatial locations denoted by $\bm{\mathrm{X}} = \{\bm{\mathrm{x}}_{i}\in D: i=1,\ldots,n_s\}$, and $n_t$ timestamps $\{t_{j}: j=1,\ldots,n_t\}$, for the forecasting tasks, our goal is to learn a function $F(\cdot)$ as our model for approximating the true mapping of historical $n_t \times n_s$ observed physical quantities to the future $n'_t \times n_s$  quantities, that is
\begin{align}
    &[u(\bm{\mathrm{X}},t_{1}), \ldots, u(\bm{\mathrm{X}},t_{n_t}) ] \notag\\
    \overset{F}{\longrightarrow}& [u(\bm{\mathrm{X}},t_{n_t+1}), \ldots, u(\bm{\mathrm{X}},t_{n_t+n'_t})].
\end{align}

\normalsize
For spatially-inductive forecasting, the learned mapping $F$ can generalize well for any spatial location $\bm{\mathrm{x}}_{\mathrm{new}} \in D$ which are unseen to the model and not included in $\bm{\mathrm{X}}$ for training the model, i.e. $\bm{\mathrm{x}}_{\mathrm{new}}\not\in\bm{\mathrm{X}}$, given its previous observations $[u(\bm{\mathrm{x}}_{\mathrm{new}},t_{1}), \ldots, u(\bm{\mathrm{x}}_{\mathrm{\mathrm{new}}},t_{n_t})]$.
For temporally-irregular forecasting, the quantities can be non-uniformly sampled, i.e. $t_{j + 1} - t_{j} \not= t_{j} - t_{j-1}$, and for other unobserved timestamps $t \in [0,T]$, the model generalizes well.

To generalize to both continuous spatial domain $D$ and temporal domain $[0,T]$, our method aims to directly model the internal mechanisms of system's dynamics by learning the spatio-temporal patterns according to PDEs with neural operators, rather than fit the discrete signals on spatial and temporal domains auto-regressively.

\section{Proposed Methods}
\subsection{Graph Kernel Encoder}
\label{sec:graphkerenc}

\begin{figure}[]
    \centering
    \includegraphics[width=1.1\linewidth]{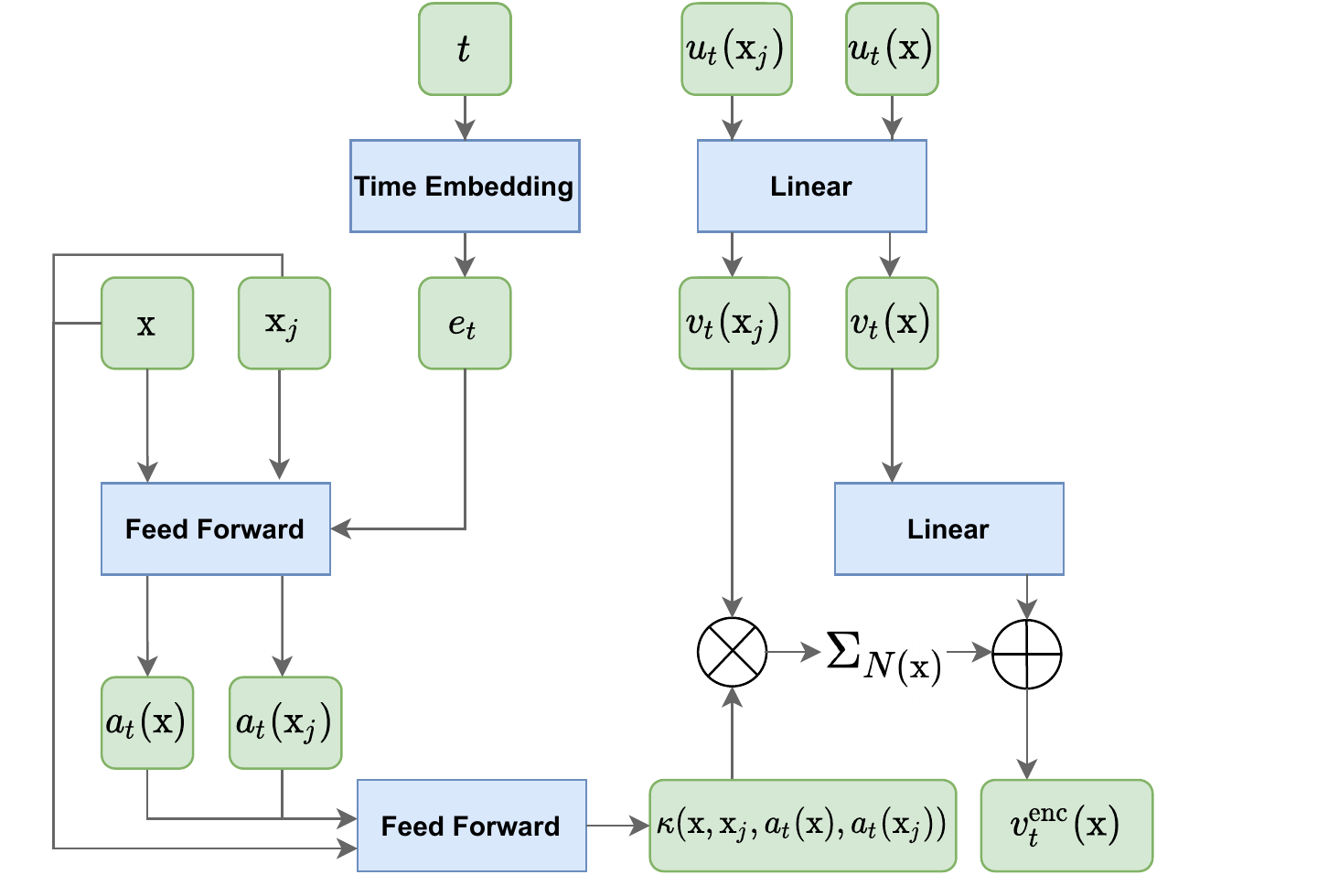}\vspace{-0.4cm}
    \caption{The workflow to update the history observations into solution representation functions. $\bm{\mathrm{x}}_j$ is in the neighborhood of $\bm{\mathrm{x}}$. $\oplus$ and $\otimes$ are element-wise add and product, and $\sum_{N(\bm{\mathrm{x}})}$ means the summation of $\bm{\mathrm{x}}$'s neighborhood.}
    \label{fig:graphkernelenc}\vspace{-0.4cm}
\end{figure}

Since the timestamps of input observations are usually fixed, we attempt to model the spatial dependency at each time $t_{j}$ in the encoder.
In this way, we assume that the underlying PDE is supposed to be of the Eq.~\ref{eq:spacediffeq} formulation. 
Therefore, the encoder of history observations of our model can be chosen as neural-operators-based in Sec.~\ref{sec:neuraloperator}, whose architecture at a single timestamp is given in Fig.~\ref{fig:graphkernelenc}.   
\vspace{-0.3cm}
\paragraph{Solution representation function.} For an observed quantity $u_t(\bm{\mathrm{x}}) = u(\bm{\mathrm{x}},t)$, we first lift $u(\bm{\mathrm{x}},t)$ into a higher dimensional representation space with a linear transformation to increase expressiveness, which can be written as $P(u(\bm{\mathrm{x}},t)) = v(\bm{\mathrm{x}},t)\in \mathbb{R}^d$, where $v(\cdot,\cdot) \in \mathcal{U}^d$ is defined as representation function. 
We assume that there exists a true solution representation function in the function space $\mathcal{U}^d$, and we aim to use $v^{\mathrm{enc}}$ for approximating it after the mapping of the encoders based on learned operator.
\vspace{-0.3cm}
\paragraph{Graph kernel operator.} To encode the history observations into representation space, we use an operator to update $v_t$ to approximate the true solution representation function.  We follow kernel operators \cite{li2020neural, li2020multipole} for parametric PDEs, which is inspired by the solution of uniformly elliptic operator and empirically proved discretization-invariant, to model the action of the integral operator written as
\begin{align}
    u_t(\bm{\mathrm{x}}) = \mathcal{F^\dag}(a_t)(\bm{\mathrm{x}}) =  \int_D G_{a_t}(\bm{\mathrm{x}}, \bm{\mathrm{y}})\left[f(\bm{\mathrm{y}},t) + \Gamma_{a_t}(\bm{\mathrm{y}})\right]d\bm{\mathrm{y}},
    \label{eq:greenformula}
\end{align}
where $G_{a_t}$ is a Newtonian potential and $\Gamma_{a_t}$ is an operator defined by appropriate sums and compositions of the modified trace and co-normal derivative operators \cite{BoundaryElement}.
Lending the Eq.~\ref{eq:greenformula} representing an integral operator as iterative architecture, it defines the operator $\mathcal{K}_{a_t}:\mathcal{U}^d\rightarrow \mathcal{U}^d$ as
\begin{align}
     (\mathcal{K}_{a_t}v_t)(\bm{\mathrm{x}}) &= \int_D \kappa_\phi(a_t(\bm{\mathrm{x}}), a_t(\bm{\mathrm{y}}), \bm{\mathrm{x}}, \bm{\mathrm{y}})v_t(\bm{\mathrm{y}})d\bm{\mathrm{y}},
\end{align}
\normalsize
where $\kappa_\phi$ with learnable parameters $\phi$ is the kernel function taking spatial locations $(\bm{\mathrm{x}}$, $\bm{\mathrm{y}})$ and values of parametric function $(a_t(\bm{\mathrm{x}})$, $a_t(\bm{\mathrm{y}}))$ as its inputs. As such, the corresponding update approximation which mimics the message passing neural network \cite{gilmer2017neural} is obtained by

\vspace{-0.3cm}
\small
\begin{align}
    &v_t^{\mathrm{enc}}(\bm{\mathrm{x}}) = (\mathcal{K}_{a_t}v_t)(\bm{\mathrm{x}}) \label{eq:encope}\\
    \approx & \sigma\left(Wv_t(\bm{\mathrm{x}}) + \frac{1}{|N(\bm{\mathrm{x}})|} \sum_{\bm{\mathrm{x}_j} \in N(\bm{\mathrm{x}})}\kappa_\phi(a_t(\bm{\mathrm{x}}), a_t(\bm{\mathrm{x}_j}), \bm{\mathrm{x}}, \bm{\mathrm{x}_j})v_t(\bm{\mathrm{x}_j})\right), \notag
\end{align}
\normalsize
where $W\in \mathbb{R}^{d\times d}$ is learnable weights, and $N(\bm{\mathrm{x}})$ is the neighborhood of $\bm{\mathrm{x}}$, which is established by an $\epsilon$-ball algorithm (See Appendix A.2.). In specific, we formulate $\kappa_\phi$ as a feed-forward neural network. By using the proposed multipole graph kernel network, the updating steps are able to capture the long-range spatial dependency between points \cite{YING2004591,li2020multipole} as well as handle the signals with \textit{spatially-irregular distribution}. 
The proposed encoder allows \textit{spatially-inductive} and \textit{temporally-irregularly forecasting}: 
When an observation $u_t(\bm{\mathrm{x}}_{\mathrm{new}})$ on an unseen spatial point $\bm{\mathrm{x}}_{\mathrm{new}}$ is obtained, the $v_t^{\mathrm{enc}}(\bm{\mathrm{x}}_{\mathrm{new}})$ can be inductively obtained according to Eq.~\ref{eq:encope} as long as its neighborhood $N(\bm{\mathrm{x}}_{\mathrm{new}})$ is known.
Moreover, for any input $u_t$, we can always obtain $v_{t}^{\mathrm{enc}}$ regardless of non-uniform intervals between timestamps.

\vspace{-0.2cm}
\paragraph{Parametric function formulation.} For classical PDE models which the kernel operators try to solve, the parametric functions are known. In contrast, the parametric functions are unknown in spatio-temporal forecasting tasks.
Thus, we need to approximate $a_t(\bm{\mathrm{x}})$ parameterized by timestamps $t$ and taking location $\bm{\mathrm{x}}$ as its input. 
We first employ trigonometric transformation based on positional encodings used in transformer language models \cite{vaswani2017attention} to embed timestamps into $d_t$-dimensional space.
\begin{align}
    \bm{e}_t = [\sin(\omega_1 t), \cos(\omega_1 t), \ldots, \sin(\omega_{d_t} t), \cos(\omega_{d_t} t)],
\end{align}
where $\{\omega_k\}_{1\leq k \leq d_t}$ are learnable parameters, and $t$ can be a vector containing time features, such as year, month, day and hour.
Then, the  parametric function is formulated as a feed-forward network taking the concatenation of time embeddings and locations as inputs, i.e. $a_t(\bm{\mathrm{x}}) = \mathrm{MLP}([\bm{\mathrm{x}}, \bm{e}_t])$. 

\subsection{Deep Operator Decoder}
\label{sec:deepoperdec}
\begin{figure}[htb]
    \centering
    \vspace{-0.4cm}
    \includegraphics[width=0.9\linewidth]{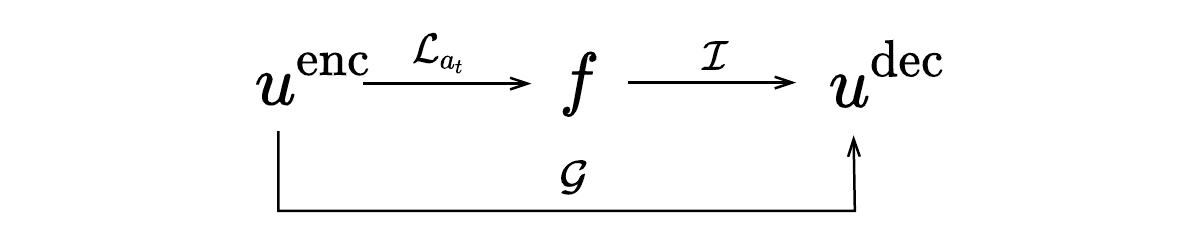}\vspace{-0.4cm}
    \caption{The construction of $\mathcal{G} = \mathcal{I}\circ\mathcal{L}_{a_t}$. To differentiate the solution function fed in the encoder and generated by the decoder, we use $u^{\mathrm{enc}}$ and $u^{\mathrm{dec}}$ to denote them.}
    \label{fig:operatormap}\vspace{-0.2cm}
\end{figure}

In the decoder which is used to generate future predictions given the past representation functions, the values of physical quantities of ground-truth are not given in the prediction process. Therefore, we cannot firstly lift $u(\bm{\mathrm{x}}, t)$ into the representation space for updating.
To address it, combined Eq.~\ref{eq:diffeq} and Eq.~\ref{eq:spacediffeq}, we write the PDE as
\begin{align}
    \frac{\partial u(\bm{\mathrm{x}},t)} {\partial t}  = (\mathcal{\bm{L}}_{a_t} u_t)(\bm{\mathrm{x}}) = f(\bm{\mathrm{x}},t),
\end{align}
which implies $u(\bm{\mathrm{x}},t) =  \int_{0}^{t}f(\bm{\mathrm{x}},\tau)d\tau = (\mathcal{G}u)(\bm{\mathrm{x}}, t)$. Let $\mathcal{I}(f)(\bm{\mathrm{x}},t) = \int_{0}^{t}f(\bm{\mathrm{x}},\tau)d\tau$, where $\mathcal{I}: \mathcal{U} \rightarrow \mathbb{R}$ is the intergal operator, and the construction of $\mathcal{G}$ is shown in Fig.~\ref{fig:operatormap}.

Inspired by this, we try to use $\mathcal{G}^\dag$ to approximate $\mathcal{G}$, to obtain the representation function solution as
\begin{align*}
    v^{\mathrm{dec}}(\bm{\mathrm{x}}, t) \approx (\mathcal{G}^\dag v^{\mathrm{enc}})(\bm{\mathrm{x}}, t).
\end{align*}
\vspace{-0.6cm}
\paragraph{Universal approximation for operator.} Following the Universal Operators in Sec.~\ref{sec:neuraloperator}, we formulate $\mathcal{G}^\dag$ according to the theorem below.
\vspace{+0.2cm}
\begin{theorem}{\rm\textbf{(Universal Approximation Theorem for Operator)} \cite{Chen1995Universal}} Suppose that $\sigma$ is a continuous non-polynomial function, $D$ is a Banach Space, $D_1 \in D, D_2 \in \mathbb{R}^{M}$ are two compact sets in $D$ and $\mathbb{R}^M$, respectively,
    $V$ is a compact set in $C(D_1)$, and $\mathcal{G}$ is a nonlinear continuous operator, which maps $V$ into $C(D_2)$.
    For any $\epsilon > 0$, there are positive integers $s,n,r$, constants $c_i^k, \xi_{ij}^{k}, \theta_i^k, \zeta_k \in \mathbb{R}$, $\bm{\eta}_k \in \mathbb{R}^m, \bm{\mathrm{p}}_j \in D_1$, $i = 1,\ldots,s; j=1,\ldots,n; k = 1,\ldots,r$, such that\\
    \small
    \begin{align}
        \left | \mathcal{G}(u)(\bm{\mathrm{q}}) - \mathcal{G}^\dag(u)(\bm{\mathrm{q}}) \right| &< \epsilon,
    \end{align}
    \normalsize
    where 
    \small
    $$
    \mathcal{G}^\dag(u)(\bm{\mathrm{q}}) = \sum_{k=1}^{r}\sum_{i=1}^s c_i^k \sigma \left( \sum_{j=1}^n\xi_{ij}^k u(\bm{\mathrm{p}}_j) + \theta_{i}^k\right)\sigma(\bm{\eta}_k \bm{\mathrm{q}} + \zeta_k)
    $$
    \normalsize
    holds for all $u\in V$ and $\bm{\mathrm{q}}\in D_2$. 
\end{theorem}
\begin{figure}[]\centering
    \hspace{-0.3cm}
    \includegraphics[width=0.9\linewidth]{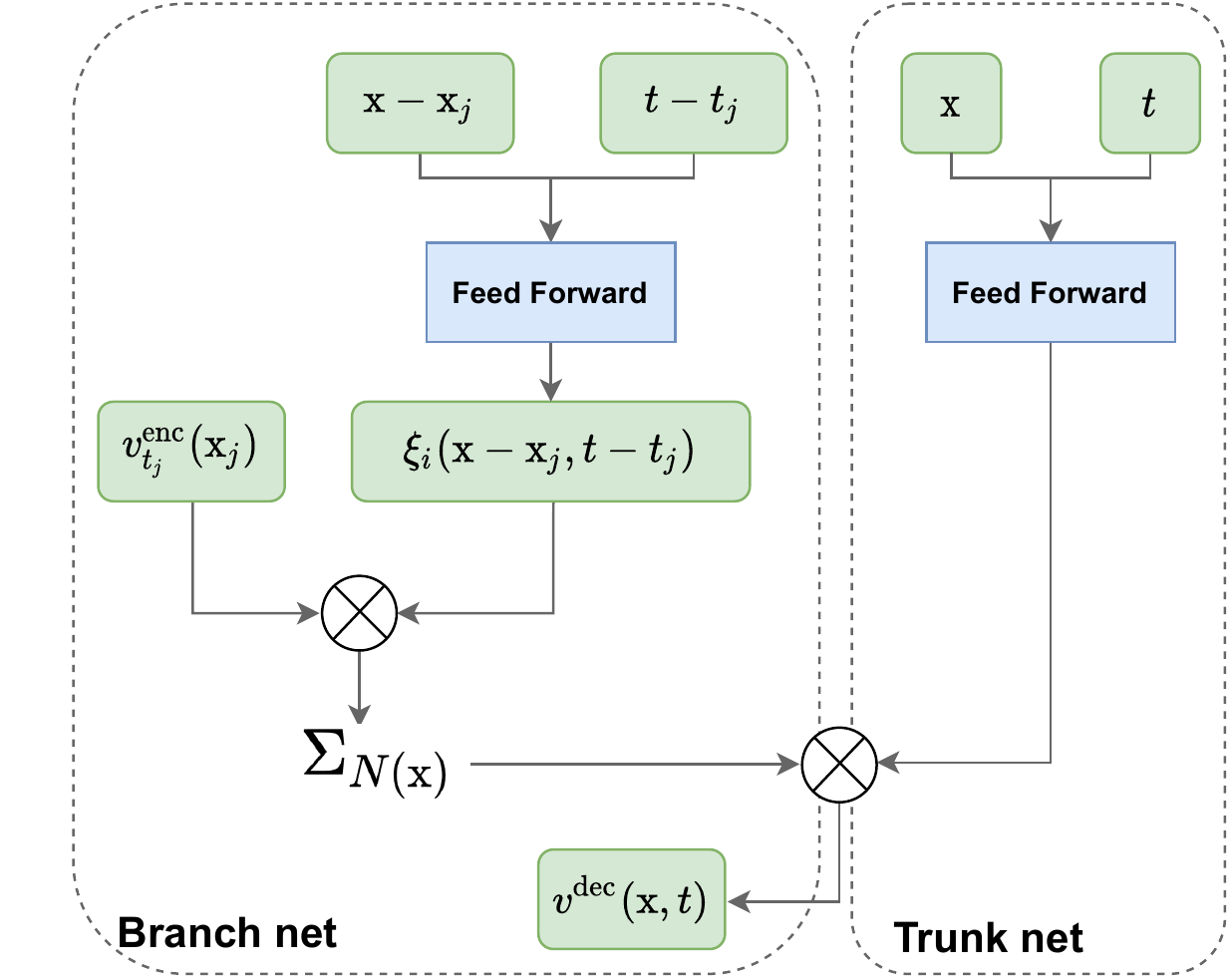}\vspace{-0.3cm}
    \caption{The workflow to approximate the solution representation functions for future prediction using past representations. $\bm{\mathrm{x}_j}$ is in the neighborhood of $\bm{\mathrm{x}}$. }
    \label{fig:unioperatordec}\vspace{-0.4cm}
\end{figure}
\normalsize
This approximation theorem indicates the potential application of neural networks to learn nonlinear
 operators from data. Thereby, we let $\bm{\mathrm{p}}_j\in\{(\bm{\mathrm{x}}_{j}, t_{j}): j = 1,\ldots, n_s\times n_t\}$, and $M = m + 1$. % where $j_s$ is the spatial sample index, and $j_t$ is the time index of history observations.
For each future timestamp $t > t_{n_t}$ for prediction, we write its representation solution function as 
{\small
\begin{align}
    &v_k^{\mathrm{dec}}(\bm{\mathrm{x}}, t) \label{eq:predrepres}\\
    =&\sum_{i=1}^d c_i^k \sigma \left( \sum_{j=1}^{n_s \times n_t}\bm{\xi}_{ij}^k v_{t_j}^{\mathrm{enc}}(\bm{\mathrm{x}}_j) + \theta_{i}^k\right)\sigma(\bm{\eta}_k[\bm{\mathrm{x}}, t] + \zeta_k), \notag
\end{align}  
}
\normalsize
where $v^{\mathrm{enc}}_{t_j}(\bm{\mathrm{x}}_j), \bm{\xi}_{ij}^k, \bm{\eta}_k \in \mathbb{R}^d$, $c_{i}^k, \theta_i^k, \zeta_k \in \mathbb{R}$ and $v^{\mathrm{dec}}_k(\bm{\mathrm{x}}, t)$ is the $k$-th component of $v^{\mathrm{dec}}(\bm{\mathrm{x}}, t) \in \mathbb{R}^d$. 
Following \textsc{DeepONet} \cite{lulu2021deeponet}, we first put the target points $(\bm{\mathrm{x}}, t)$ into a feed-forward network called Trunk net,
{\small
\begin{align}
    \mathrm{MLP}_{ik}^{\mathrm{Tr}}(\bm{\mathrm{x}}, t) = \sigma(\bm{\eta}_k[\bm{\mathrm{x}}, t] + \zeta_k),
\end{align}
}
\normalsize and put all the obtained past representations $\{v^{\mathrm{enc}}_{t_j}(\bm{\mathrm{x}}_j)\}$ into the other network called Branch net, as
{\small
\begin{align}
    \mathrm{MLP}_{ik}^{\mathrm{Br}}(\bm{\mathrm{x}}_j,t_j) = \sigma \left( \sum_{j=1}^{n_s \times n_t}\bm{\xi}_{ij}^k v_{t_j}^{\mathrm{enc}}(\bm{\mathrm{x}}_j) + \theta_{i}^k\right). \label{eq:brdec}
\end{align}}
\normalsize
By multiplying the two outputs of $\mathrm{MLP}^{\mathrm{Tr}}(\bm{\mathrm{x}}, t)$ and $\mathrm{MLP}^{\mathrm{Br}}(\bm{\mathrm{x}}, t)$ with a linear transformation whose parameter is $\{c_i^k\}$ stacked after, we obtain the representation function $v^{\mathrm{dec}}(\bm{\mathrm{x}}, t) = (v_1^{\mathrm{dec}}(\bm{\mathrm{x}}, t), \ldots, v_d^{\mathrm{dec}}(\bm{\mathrm{x}}, t))$ for prediction. 
\vspace{-0.25cm}
\normalsize
\paragraph{Revised flexible decoder.} 
The construction of the Branch-Trunk decoder according to Eq.~\ref{eq:predrepres} is time-continuous, so it is able to solve the \textit{temporally-irregular forecasting} theoretically.
However, this establishment does not allow inductive forecasting. The reason for it is that:

When a sequence of observations on an unseen spatial point $\bm{\mathrm{x}}_{\mathrm{new}}$ is obtained, the Branch Net in Eq.~\ref{eq:brdec} of the decoder does not permit $\{v^{\mathrm{enc}}_{t_j}(\bm{\mathrm{x}}_{\mathrm{new}})\}$ which is inductively obtained by the encoder to be added to $\{v^{\mathrm{enc}}_{t_j}(\bm{\mathrm{X}})\}$ to compute $\{v^{\mathrm{dec}}(\bm{\mathrm{x}}_{\mathrm{new}},t)\}$, because the number of parameter $\{\bm{\xi}_{ij}^k\}$ for index $j$ is fixed as $n_s \times n_t$, which is inextensible.

To address it, we revise the Branch net as 
\small
\begin{align}
    &\mathrm{AGG}_{ik}^{\mathrm{Br}}(\bm{\mathrm{x}},\bm{\mathrm{x}}_j, t, t_j)  \label{eq:revdec}\\
    =& \sigma\left( \frac{1}{|N(\bm{\mathrm{x}})|}\sum_{\substack{\bm{\mathrm{x}}_j \in N(\bm{\mathrm{x}}) \\ 1\leq j\leq n_s \times n_t}}\xi_{i}^k(\bm{\mathrm{x}}- \bm{\mathrm{x}}_j, t-t_j) v_{t_j}^{\mathrm{enc}}(\bm{\mathrm{x}}_j) + \theta_{i}^k\right).\notag
\end{align}
\normalsize
We use `$\mathrm{AGG}^{\mathrm{Br}}$' to denote the revised Branch net, which means that the representation functions are updated by aggregating messages from their neighbors' previous representation functions, where the message-passing weights are inductively obtained by a feed-forward network $\xi_{i}^{k}(\cdot, \cdot)$ with relative spatial location and time as inputs.
Fig.~\ref{fig:unioperatordec} gives the workflows of our decoder model. In this way, when a sequence of history observations is added, the Branch's output can still be flexibly obtained as long as the neighborhood is known. 

In addition, an auxiliary advantage brought by Eq. \ref{eq:revdec} is that the potential explosion of model's parameter number can be avoided. In Eq.~\ref{eq:brdec}, when the number of observed points are large, the computational complexity may explore. As $n_s$ increases by 1, the parameter number will increase by $\mathcal{O}(n_t\times d \times d)$, complexity of which may be unaffordable. The revised decoder's parameter number will not change as the number of spatial sample points increases.
\vspace{-0.25cm}

\normalsize
\subsection{Loss function}
For training, after getting $\{v_t^{\mathrm{enc}}(\bm{\mathrm{X}}): t=t_{1},\ldots,t_{n_t}\}$ as the values of representation function obtained by encoder, and
$\{v^{\mathrm{dec}}(\bm{\mathrm{X}},t): t=t_{n_t + 1},\ldots,t_{n_t+n'_t}\}$ as the values of representation function obtained by decoder, where $\bm{\mathrm{X}} = \{\bm{\mathrm{x}}_{i}\in D: i=1,\ldots,n_s\}$ are the spatial locations of sample points as shown in Sec.~\ref{sec:spatemp}, 
we use a linear projector $P'$ to map them back to $\mathcal{U}$, and use mean absolute error (MAE) to construct the loss function, which reads
\begin{align}
  L =& \frac{\alpha}{n_t n_s}\sum_{t=t_{1}}^{t_{n_t}}\lVert P'(v_t^{\mathrm{enc}}(\bm{\mathrm{X}})) - u(\bm{\mathrm{X}},t) \rVert_1  \\
     & + \frac{1}{n'_t n_s}\sum_{t=t_{n_t+1}}^{t_{n_t+n'_t}}\lVert P'(v_t^{\mathrm{dec}}(\bm{\mathrm{X}})) - u(\bm{\mathrm{X}},t) \rVert_1,\notag
\end{align}
where the first term we call the reconstruction loss, to restrict the encoder and projector to preserve information from inputs, with $\alpha$ `\textit{the weight of reconstruction loss}',
and the second term we call the prediction loss, to evaluate the prediction performance of the model. 

\section{Related Works}
\paragraph{Graph Spatio-temporal networks.} Spatio-temporal forecasting networks are mostly graph-based thanks to their ability to learn representations of spatially-irregular distributed signals, such as traffic flows recorded by sensors \citep{yu2018STGCN,li2018diffusion,guo2019attention, bai2020adaptive,Zhao2020TGCN,li2021dynamic}. 
These works usually regard signals' location as nodes, and establish graphs to describe the nodes' dependency according to their spatial distance. 
Also, some works learn the adjacency relations of nodes without the prior graph structure, by using attention mechanisms \citep{yu2018STGCN,li2018diffusion,Zhao2020TGCN}, graph structure learning techniques \cite{shang2021discrete}, node embeddings \cite{bai2020adaptive} and so on.
However, these methods are constructed for transductive tasks on discrete nodes. Recently, a spatio-temporal meteorological forecasting methods for physical quantities located on continuous space (earth sphere) is proposed \cite{lin2021conditional}, but it still aims to `fit' the signals on discrete nodes.
In comparison, our model aims to both handle irregularly distributied signals and allow spatially-inductive forecasting by employing multipole graph neural operators \cite{li2020multipole}.

\paragraph{PDE-driven spatio-temporal networks.} An increasing number of works combining spatio-temporal networks with differential equations have been produced in recent years.
A series of them are established for data with regular grids, such as videos \citep{guen2020disentangling, dona2021pdedriven}. These methods are usually constructed on (2D-image-)convolution neural networks as the basic architecture of spatial encoders, while our model aims to handle spatially-irregular distributed points, and thus is graph-based.
Another stream of PDE-driven networks follow Neural ODE \cite{chen2019neural} for modeling dynamical systems \citep{yd2019ode2vae, rubanova2019latent}. Because of the high computational cost of integral estimation and backward propagation, these models are not feasible for long-term forecasting. We also establish a version of ODE-based model (Appendix B.) and conduct experiment illustrating its infeasible time cost.

\paragraph{Neural operators.}  A line of neural networks have been designed to model one specific instance of the PDEs \citep{e2017deep,bar2019unsupervised,smith2020eikonet,shaowu2020,Raissi2020sceience} with prior physical knowledge. For example, structural priors and constraints are combined for fluid prediction \cite{tompson2017accelerating,Raissi2020sceience}, and Hamiltonian mechanics are used to construct non-regression losses \cite{greydanus2019hamiltonian,chen2020symplectic,toth2020hamiltonian} to learn basic laws of physics.
In contrast, our model aims to be applicable to general spatio-temporal forecasting. Thus, we turn to neural operators \citep{lulu2021deeponet,bhattacharya2021model,li2020multipole,li2021fourier, nelsen2021random} for solutions, which require no knowledge of the underlying PDE, only data. Besides, their learned network parameters can be generalized with different discretizations of points. In this way, we conjecture the operator can capture the implicit mechanisms of the dynamics, and thus employ it as the spatial fundamental modules of our model.

\section{Experiments}
\label{sec:experiment}
In the experiment part, we aim to figure out the following three questions:
\vspace{-0.3cm}
\begin{itemize}
    \item[\textbf{\textit{Q}.1}] How the proposed \textsc{STONet} performs: Does it achieve improved performance on forecasting tasks of continuous spatial domain, compared with other graph-based spatio-temporal neural networks?
    \vspace{-0.2cm}
    \item[\textbf{\textit{Q}.2}] Whether \textsc{STONet} allows spatially-inductive forecasting: Does the trained model achieve comparable accuracy of prediction on unseen spatial points, given their previous observations?
    \vspace{-0.2cm}
    \item[\textbf{\textit{Q}.3}] Whether \textsc{STONet} allows temporally-irregular forecasting: Does the model give accurate predictions, even when the labels of future predictions are sampled irregularly, or missing with a ratio?
\end{itemize}
\subsection{Experiment Setup} 
\paragraph{Protocol.} Because our method is graph-based to handle the spatially-irregular signals or physical quantities, we choose 8 methods which are all graph-based spatio-temporal models for performance comparison. Table.~\ref{tab:modelcomparision} and Appendix C.1. give descriptions of the methods, where  \textit{`Spatial'} and \textit{`Temporal'}  represent the  modules to capture spatial dependency and temporal dynamics. Our conclusion is based on the implementation of \textsc{Torch-Geomtemp} \cite{rozemberczki2021pytorch} and each methods' open source code.
Three widely used metrics - Mean Absolute Error
(MAE), Root Mean Square Error (RMSE), and Mean Absolute Percentage Error (MAPE) are deployed to measure the performance. 
The reported mean and standard deviation of metrics are obtain through 5 independent experiments with different random seeds.
All the models for comparison are trained with target function of MAE and optimized by Adam optimizer. 
The hyper-parameters are chosen through a carefully tuning on the validation set. In implementation, we stack several decoders for different terms of forecasting in \textsc{STONet} as analyzed in Sec.~\ref{sec:ablation} and Appendix C.5.
\begin{table}[]
    \centering
    \caption{Comparison of different spatio-temporal methods}\label{tab:modelcomparision}
    \resizebox{1.05\columnwidth}{!}{
    \begin{tabular}{llc}
        \toprule
    Methods & Spatial     & Temporal  \\
    \midrule
    \textsc{TGCN} \cite{Zhao2020TGCN}   & Vanilla GCN \cite{kipf2017semisupervised}            & GRU                      \\
    \textsc{STGCN} \cite{yu2018STGCN}   & Vanilla GCN \cite{kipf2017semisupervised}           & 1D Conv                   \\
    \textsc{MSTGCN} \cite{Guo2019ASTGCN}  & ChebConv \cite{defferrard2017convolutional}             & 1D Conv                    \\
    \textsc{ASTGCN} \cite{Guo2019ASTGCN}   & GAT \cite{veli2018graph}                   & Attention              \\
    \textsc{GCGRU} \cite{seo2016gcgru}   & ChebConv  \cite{defferrard2017convolutional}         & GRU                        \\
    \textsc{DCRNN} \cite{li2018diffusion}   & DiffConv \cite{atwood2016diffusionconvolutional}            & GRU                          \\
    \textsc{AGCRN} \cite{bai2020adaptive}  & Node Similarity \cite{bai2020adaptive}       & GRU                    \\
    \textsc{CLCRN} \cite{lin2021conditional}  & CondLocalConv \cite{lin2021conditional}       & GRU                   \\
\bottomrule    
\end{tabular}\vspace*{-3em}
}%
\end{table}

\paragraph{Dataset.} We conduct an evaluation on 6 datasets, whose forecasting target signals are all spatially-continuous physical quantities, including \texttt{Wave} \cite{saha2021physicsincorporated}, \texttt{Solar energy} \cite{lai2018modeling}, \texttt{Temperature}, \texttt{Humidity}, \texttt{Cloud cover} and \texttt{Wind component} \cite{rasp2020weatherbench}.
All the evaluation tasks are 12-to-12 auto-regressive forecasting, where the input and forecasting sequence length are all set as 12.
The spatial domain includes both 2D plane and 2D sphere (earth's surface). Dataset descriptions are shown in Table.~\ref{tab:dataset}. For \textsc{CLCRN} \cite{lin2021conditional}, it is established for dataset whose spatial domain is sphere, so we do not compare its performance on planar dataset. To differentiate from the time point, we call the fixed spatial points as nodes, because all the methods are graph-based.
\begin{table}[htb]\vspace{-1em}
    \caption{Dataset discription}\label{tab:dataset}
    \resizebox{1.00\columnwidth}{!}{
        \begin{tabular}{c|ccc}
            \toprule
            Datasets           & \texttt{\small Wave} & \texttt{\small Solar Energy} & \texttt{\small Temperature}            \\
            \midrule
            Spatial domain           &2D plane        &2D plane   &2D sphere                 \\
            Dimension          &1            &1       & 1                          \\
            \# of nodes        & 512          & 137          & 2048                      \\
            Granularity        & 0.001s       & 5 min          & 1h        \\  

            \toprule
            Datasets        & \texttt{\small Humidity} & \texttt{\small Cloud cover} & \texttt{\small Wind component}              \\
            \midrule
            Spatial domain               &2D sphere       &2D sphere    &2D sphere            \\
            Dimension              & 1           & 1        & 2                 \\
            \# of nodes               & 2048        & 2048     & 2048              \\
            Granularity          & 1h      & 1h  & 1h    \\
            \bottomrule    
        \end{tabular}}
    \end{table}\vspace{-2em}

\subsection{Performance Comparison}
We first conduct experiments on the datasets with 8 baselines and our \textsc{STONet}. Because MAPE is of great difference among methods and hard to agree on an order of magnitude due to its incompatible units, we show it in Appendix C.2.
\begin{table*}[ht]
    \centering
    \caption{Forecasting results on different datasets. Results in \textbf{bold} are the top-1  performance, and results with \underline{underlines} are the second. `\textit{Improvements}' is the percentage of top-1 over the second.} \label{tab:performcomparision}
    \resizebox{1.60\columnwidth}{!}{
    \begin{tabular}{crrrrrr}
        \toprule
    \multicolumn{1}{c}{}                          & \multicolumn{1}{c}{MAE} & \multicolumn{1}{c}{RMSE} & \multicolumn{1}{c}{MAE} & \multicolumn{1}{c}{RMSE} & \multicolumn{1}{c}{MAE} & \multicolumn{1}{c}{RMSE} \\
    \cmidrule(lr){2-3} \cmidrule(lr){4-5} \cmidrule(lr){6-7}
     & \multicolumn{2}{c}{\texttt{Wave} ({\scriptsize$\times 10^{-4}$})}                           & \multicolumn{2}{c}{\texttt{Solar Energy} ({\scriptsize MW})}                   & \multicolumn{2}{c}{\texttt{Temperature} ({\scriptsize K})}                    \\
     \midrule
     \textsc{TGCN}                                          & 3.7619{\scriptsize±0.0144}                         & 5.8999{\scriptsize±0.0485}                  & 2.0036{\scriptsize±0.0423}                          & 4.6218{\scriptsize±0.3820}                           & 3.8638{\scriptsize ±0.0970}                   & 5.8554{\scriptsize±0.1432}            \\
    \textsc{STGCN}                                         & 4.1202{\scriptsize±0.1800}                          & 6.4805{\scriptsize±0.4165}                           & 1.4048{\scriptsize±0.0184}                          & 4.1451{\scriptsize±0.0249}                          & 4.3525{\scriptsize±1.0442}                    & 6.8600{\scriptsize±1.1233}            \\
    \textsc{MSTGCN}                                        & 6.0639{\scriptsize±0.2664}                          & 7.8891{\scriptsize±0.5628}                           & 1.9224{\scriptsize±0.0251}                          & 4.0406{\scriptsize±0.0471}                          & 1.2199{\scriptsize±0.0058}           & 1.9203{\scriptsize±0.0093}            \\
    \textsc{ASTGCN}                                        & 5.3150{\scriptsize±0.2012}                          & 7.1699{\scriptsize±0.2045}                           & 1.9834{\scriptsize±0.0054}                          & 4.1312{\scriptsize±0.0095}                          & 1.4896{\scriptsize±0.0130}                    & 2.4622{\scriptsize±0.0023}            \\
    \textsc{GCGRU}                                         & \underline{3.5617}{\scriptsize±0.4817}             & \underline{5.6927}{\scriptsize±0.4181}                  & \underline{1.0661}{\scriptsize±0.0867}                 & \underline{2.6303}{\scriptsize±0.1197}                 & 1.3256{\scriptsize±0.1499}                    & 2.1721{\scriptsize±0.1945}            \\
    \textsc{DCRNN}                                         & 3.7338{\scriptsize±0.1167}                         & 5.8157{\scriptsize±0.0859}                           & 1.1031{\scriptsize±0.1140}                 & 2.7474{\scriptsize±0.2229}                          & 1.3232{\scriptsize±0.0864}                    & 2.1874{\scriptsize±0.1227}            \\
    \textsc{AGCRN}                                         & 4.2048{\scriptsize±0.1161}                          & 6.1559{\scriptsize±0.0776}                           & 1.1845{\scriptsize±0.0902}                          & 2.6823{\scriptsize±0.1261}                 & 1.2551{\scriptsize±0.0080}                    & 1.9314{\scriptsize±0.0219}            \\
    \textsc{CLCRN}                                         & \multicolumn{1}{c}{-}                                                     & \multicolumn{1}{c}{-}                                                      & \multicolumn{1}{c}{-}                                                     & \multicolumn{1}{c}{-}                                                     & \underline{1.1688}{\scriptsize±0.0457}           & \underline{1.8825}{\scriptsize±0.1509}            \\
    \textsc{\textbf{STONet}}                               & \textsc{\textbf{3.1959}}{\scriptsize±0.0722}                 & \textsc{\textbf{5.2201}}{\scriptsize±0.0743}                  & \textsc{\textbf{0.8699}}{\scriptsize±0.0399}                 & \textsc{\textbf{2.1533}}{\scriptsize±0.0563}                 & \textsc{\textbf{0.8972}}{\scriptsize±0.0230}           & \textsc{\textbf{1.4963}}{\scriptsize±0.0422}   \\        
    \midrule
    \small Improvements                                    &\textbf{10.2703\%}                                     &\textbf{8.3019\%}&\textbf{18.4035\%}&\textbf{18.1348\%} &\textbf{23.2375\%}  &\textbf{20.5153\%} \\

    \bottomrule
     & \multicolumn{2}{c}{\texttt{Humidity} ({\scriptsize $\%\times 10$})}                       & \multicolumn{2}{c}{\texttt{Cloud Cover} ({\scriptsize $\%\times 10^{-1}$})}                    & \multicolumn{2}{c}{\texttt{Wind Component} ({\scriptsize $\mathrm{ms}^{-1}$})}                 \\
    \midrule
    \textsc{TGCN}                                          & 1.4700{\scriptsize±0.0295}                    & 2.1066{\scriptsize±0.0551}                     & 2.3934{\scriptsize±0.0216}                      & 3.6512{\scriptsize±0.0223}                      & 4.1747{\scriptsize±0.0324}                      & 5.6730{\scriptsize±0.0412}            \\
    \textsc{STGCN}                                         & 0.7975{\scriptsize±0.2378}                    & 1.1109{\scriptsize±0.2913}                     & 2.0197{\scriptsize±0.0392}                      & 2.9542{\scriptsize±0.0542}                      & 3.6477{\scriptsize±0.0000}                      & 4.8146{\scriptsize±0.0003}            \\
    \textsc{MSTGCN}                                        & 0.6093{\scriptsize±0.0012}                    & 0.8684{\scriptsize±0.0019}                     & 1.8732{\scriptsize±0.0010}                      & 2.8629{\scriptsize±0.0073}                      & 1.9440{\scriptsize±0.0150}                      & 2.9111{\scriptsize±0.0292}            \\
    \textsc{ASTGCN}                                        & 0.7288{\scriptsize±0.0229}                    & 1.0471{\scriptsize±0.0402}                     & 1.9936{\scriptsize±0.0002}                      & 2.9576{\scriptsize±0.0007}                      & 2.0889{\scriptsize±0.0006}                      & 3.1356{\scriptsize±0.0012}            \\
    \textsc{GCGRU}                                         & 0.5007{\scriptsize±0.0002}                    & 0.7891{\scriptsize±0.0006}                     & 1.5925{\scriptsize±0.0023}             & 2.5576{\scriptsize±0.0116}                      & 1.4116{\scriptsize±0.0057}                         & 2.2931{\scriptsize±0.0047}            \\
    \textsc{DCRNN}                                         & 0.5046{\scriptsize±0.0011}                    & 0.7956{\scriptsize±0.0033}                     & 1.5938{\scriptsize±0.0021}                      & 2.5412{\scriptsize±0.0044}             & 1.4321{\scriptsize±0.0019}                      & 2.3364{\scriptsize±0.0055}            \\
    \textsc{AGCRN}                                         & 0.5759{\scriptsize±0.1632}                    & 0.8549{\scriptsize±0.2025}                     & 1.7501{\scriptsize±0.1467}                      & 2.7585{\scriptsize±0.1694}                      & 2.4194{\scriptsize±0.1149}                      & 3.4171{\scriptsize±0.1127}            \\
    \textsc{CLCRN}                                         & \underline{0.4531}{\scriptsize±0.0065}           & \underline{0.7078}{\scriptsize±0.0146}            & \textsc{\textbf{1.4906}}{\scriptsize±0.0037}             & \underline{2.4559}{\scriptsize±0.0027}             & \underline{1.3260}{\scriptsize±0.0483}             & \underline{2.1292}{\scriptsize±0.0733}            \\
    \textsc{\textbf{STONet}}                               & \textsc{\textbf{0.4273}}{\scriptsize±0.0256}           & \textsc{\textbf{0.6584}}{\scriptsize±0.0287}            & \underline{1.4933}{\scriptsize±0.0030}             & \textsc{\textbf{2.4142}}{\scriptsize±0.0016}             & \textsc{\textbf{1.2192}}{\scriptsize±0.0064}             & \textsc{\textbf{1.9774}}{\scriptsize±0.0122}            \\
    \midrule
    \small Improvements                                           &\textbf{5.6841\%}  &\textbf{6.9794\%} & \multicolumn{1}{c}{\textbf{-0.1811\%}} & \textbf{1.7794\%}&\textbf{8.0543\%}&\textbf{7.1294\%} \\
    \bottomrule    
\end{tabular}}\vspace{-1em}
\end{table*}

From Table.~\ref{tab:performcomparision}, it can be concluded that
\vspace{-0.2cm}
\begin{itemize}
    \item Due to the high expressivity of \textsc{STONet}, it outperforms other models for comparison with a large margin on all but the smallest of the benchmark datasets such as \texttt{Cloud Cover}.
    \item Because most of the compared methods are established for traffic forecasting on discrete nodes, they show a decrease in performance for continuous physical quantity forecasting tasks. The difference between two tasks is analyzed in previous works \cite{lin2021conditional}. 
\end{itemize}
\subsection{Spatially Inductive Evaluation}
The second part is to figure out \textbf{\textit{Q}.2} : whether the proposed \textsc{STONet} allows \textit{spatially-inductive forecasting}.  
We preprocess the datasets to ensure that there are unseen the spatial nodes in the datasets for models to be trained with. For example, 
for the four weather datasets, we first downsample the resolution to $16\times32$, with $512$ nodes used for training, and randomly choose another $512$ different nodes for inductive evaluation.   
Other details on preprocessing datasets for this task are described in Appendix C.3.
% For \texttt{Wave}, The nodes are originally sampled from $64 \times 64$ images, so we randomly choose another $512$ nodes to evaluate models' spatially-inductive forecasting ability. 
% \texttt{Solar Energy}: Since it is a dataset with low-resolution, we randomly choose $108$ nodes for training, and the rest $29$ nodes are used for inductive evaluation.
%  Four weather datasets: We first downsample the resolution to $16\times32$, with $512$ nodes used for training, and randomly choose another $512$ different nodes for inductive evaluation.   

\begin{figure*}[ht]
    \centering
            \subfigure[MAE and Deviation on Temperature.]{ \label{fig:inducttemperature}
                \includegraphics[height=0.257\linewidth]{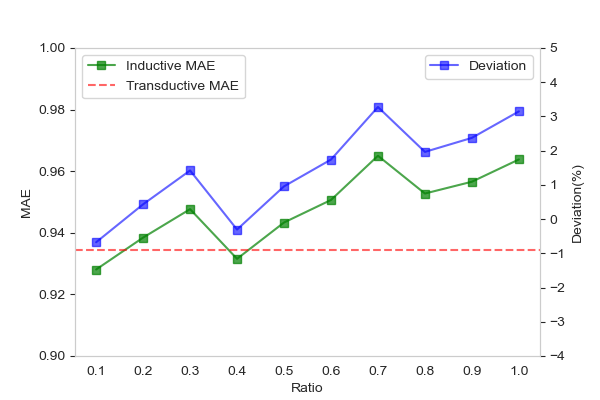}}\hspace{0mm}
            \subfigure[MAE and Deviation on Humidity.]{ \label{fig:inducthumidity}
                \includegraphics[width=0.393\linewidth]{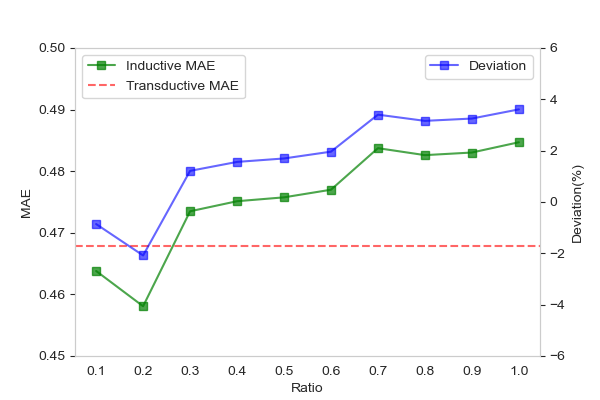}}\hspace{0mm}
            \subfigure[RMSE]{ \label{fig:inductrmse}
                \includegraphics[width=0.19\linewidth]{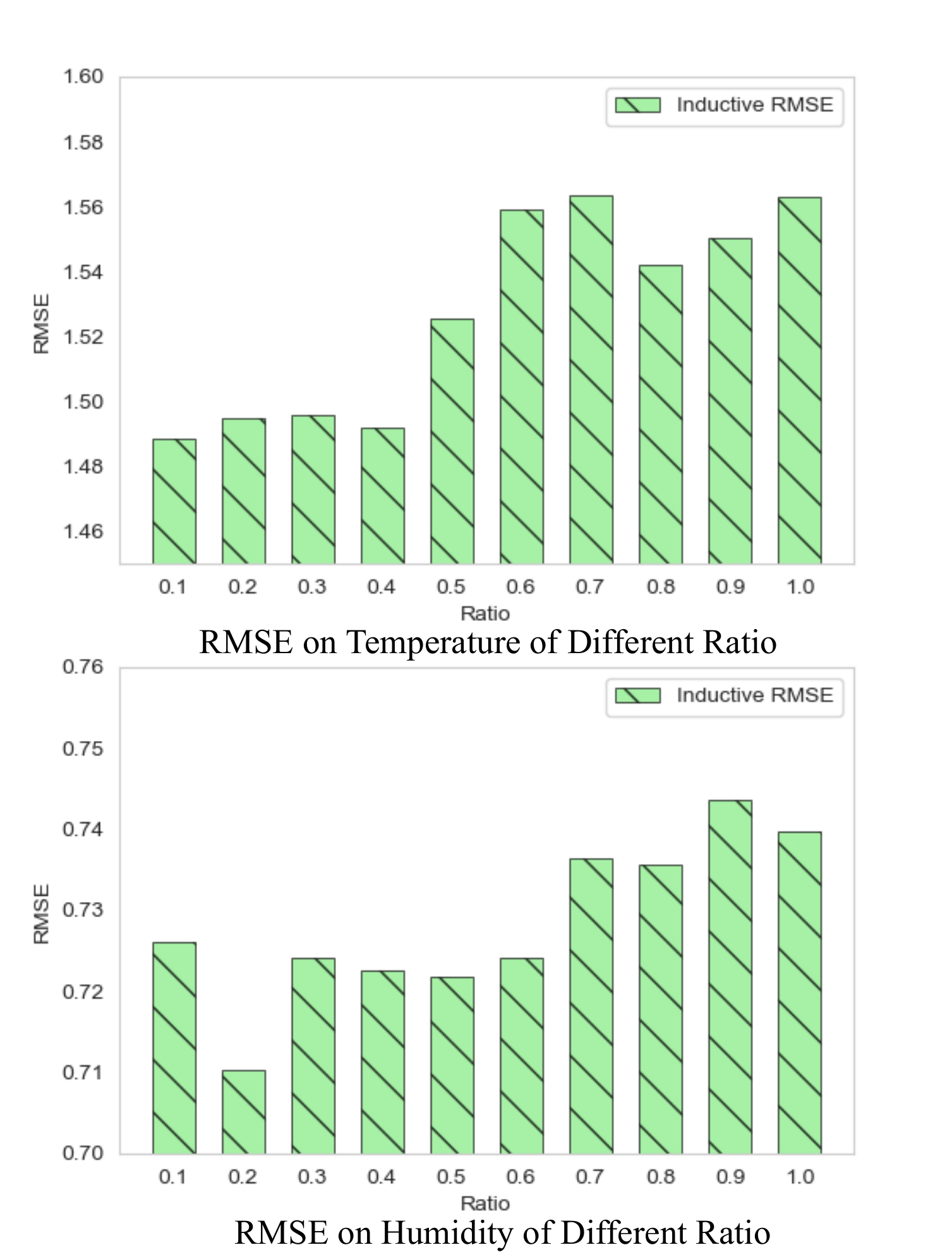}} \vspace{-0.3cm}
    \caption{The change of MAE, Deviation and RMSE on different ratio of inductive nodes number on Dataset of \texttt{Humidity} and \texttt{Temperature}. For other results, see Appendix C.3.}\vspace{-0.4cm} \label{fig:inductive}
\end{figure*}

Our results of performance on original nodes (for transductive forecasting) and unseen nodes (for inductive forecasting) are shown in Table.~\ref{tab:inductive}, where \textit{`Deviation'} is calculated by $100 \% \times \frac{\mathrm{MAE}_{\mathrm{ind}} - \mathrm{MAE}_{\mathrm{tran}} }{\mathrm{MAE}_{\mathrm{tran}}}$. Fig.~\ref{fig:inductive} shows the performance of inductive tasks on unseen nodes with different ratios, where $Ratio = \frac{\#\{\textrm{unseen node}\}}{\# \{\textrm{original node}\}} $, on several datasets. Results on other datasets are in Appendix C.3. We conclude that
\begin{table}[ht]
    \vspace{-0.3cm}
    \centering
    \caption{Evaluation results  on transductive and inductive tasks.} \label{tab:inductive}
    \resizebox{1.02\columnwidth}{!}{
        \begin{tabular}{c|cccc}
            \toprule
            Metric                & \text{Tasks}  & \texttt{Wave}          & \texttt{Solar Energy}  & \texttt{Temperature}   \\
            \midrule
            \multirow{2}{*}{\rotatebox{0}{MAE}}  & Trans. & \textsc{3.1959}{\scriptsize±0.0722}  & \textsc{1.1448}{\scriptsize±0.0206} & \textsc{0.9344}{\scriptsize±0.0252} \\
                                  & Induc. & \textsc{3.1719}{\scriptsize±0.3624} & \textsc{1.1799}{\scriptsize±0.0373} & \textsc{0.9638}{\scriptsize±0.0371} \\
            \midrule
            \multirow{2}{*}{\rotatebox{0}{RMSE}} & Trans. & \textsc{5.2586}{\scriptsize±0.0743}  & \textsc{2.4654}{\scriptsize±0.0454} & \textsc{1.5250}{\scriptsize±0.0455} \\
                                  & Induc. & \textsc{5.3794}{\scriptsize±0.2933} & \textsc{2.5028}{\scriptsize±0.0427} & \textsc{1.5631}{\scriptsize±0.0594}\\
            \midrule
            \scriptsize Deviation& &\textsc{-0.7509\%} &\textsc{+3.0660\%}&\textsc{+3.1464\%}  \\
            \bottomrule
            \toprule
            Metric                & \text{Tasks}   & \texttt{Humidity}          & \texttt{Cloud Cover}  & \texttt{Wind Component}   \\
            \midrule
            \multirow{2}{*}{\rotatebox{0}{MAE}}  & Trans. & \textsc{0.4678}{\scriptsize±0.0167} & \textsc{1.5874}{\scriptsize±0.0028} & \textsc{1.3124}{\scriptsize±0.0149} \\
                                                 & Induc. & \textsc{0.4847}{\scriptsize±0.0248} & \textsc{1.6024}{\scriptsize±0.0101} & \textsc{1.3330}{\scriptsize±0.0154} \\
            \midrule
            \multirow{2}{*}{\rotatebox{0}{RMSE}} & Trans. & \textsc{0.7075}{\scriptsize±0.0150} & \textsc{2.5223}{\scriptsize±0.0039} & \textsc{2.1588}{\scriptsize±0.0375} \\
                                                 & Induc. & \textsc{0.7398}{\scriptsize±0.0132} & \textsc{2.5213}{\scriptsize±0.0183} & \textsc{2.1459}{\scriptsize±0.0462}\\
            \midrule
            \scriptsize Deviation& &\textsc{+3.6082\%} & \textsc{+0.9449\%}   &\textsc{+1.5696\%}  \\
            \bottomrule
            \end{tabular}
    }
    \vspace{-0.3cm}
    \end{table}
%  For $\texttt{Solar Energy}$, the limited total node number does not permit the evaluation of different ratio, so we does not show it, because when the nodes for training equals to $67$, the model cannot event fit the signal well.
\vspace{-0.3cm}
\begin{itemize}
    \item The overall performance on transductive tasks shows decrease except in \texttt{Wave}, because downsampling is conducted in the other five datasets, causing the loss of spatial information.
    \vspace{-0.2cm}
    \item The evaluation on unseen nodes usually shows a tiny decrease compared with the nodes for training, but still achieves a competitive accuracy with the `\textit{Deviation}' less than $4\%$.
    \vspace{-0.2cm}
    \item As the ratio of unseen points increases, the performance on them usually shows a trend of decrease with some fluctuation. An explanation of it is that the message-passing patterns of \textsc{STONet} entangle intricately, and could possibly be disturbed when a large number of unseen nodes are included for forecasting.  
\end{itemize}
\vspace{-0.2cm}
To sum up, empirical studies illustrate that the inductive forecasting task can be well-solved by our model when the number of unseen nodes is not extremely large, since the deviation is smaller than $4\%$ in different $Ratio$s. 

\subsection{Temporally Irregular Evaluation}
To demonstrate that our model can solve the  \textbf{\textit{Q}.3}, we conduct experiments in one of the \textit{temporally-irregular forecasting} scenarios -- data with missing values, since all the evaluated datasets are all uniformly sampled in temporal domains. % In the following experiments, the labels of output timestamps are removed with different ratios.
In detail, the previous setting is all based on a $n_{t}$-to-$n'_{t}$ forecasting task, where $n_{t} = n'_{t} = 12$, while in this part, we randomly remove labels of different sequences of data at different timestamps with a ratio, leading to the non-uniformity of time-interval.
We set $n_{t'} = \#\{\textrm{output timestamps}\}$, where $n_{t'} \leq 12$. For example, if we randomly remove spatial snapshots at $2$ timestamps, we think that it is equivalent to the missing data scenario where the missing ratio equals $\frac{2}{12}$, and $n_{t'} = 10$.  
\begin{figure}[htb]\vspace{-0.3cm}
    \centering
            \subfigure[Cloud Cover.]{ \label{fig:irregularcloud}
                \includegraphics[height=0.322\columnwidth]{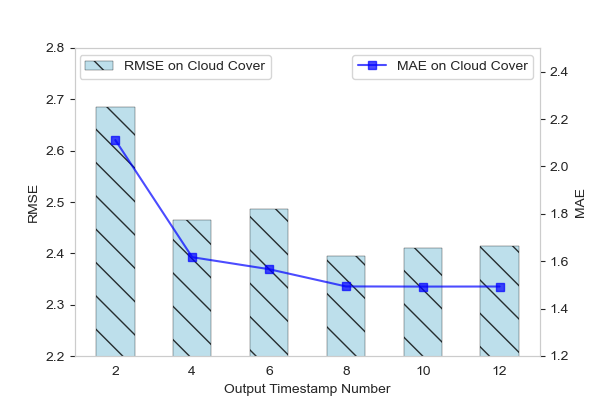}}\hspace{0mm}
            \subfigure[Wind Component.]{ \label{fig:irregularwind}
                \includegraphics[width=0.484\columnwidth]{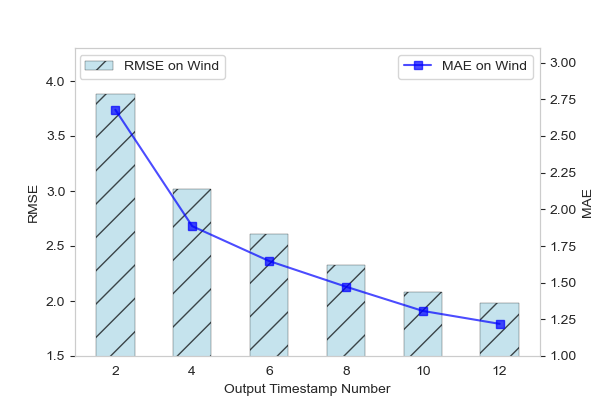}}\hspace{0mm}\vspace{-0.3cm}
    \caption{The change of MAE and RMSE on different output timestamp numbers ($n_{t'}$) on Dataset of \texttt{Cloud Cover} and \texttt{Wind Component}. For other results, see Appendix C.4.}\vspace{-0.4cm} \label{fig:irregulartime}
    \end{figure}
    
Fig.~\ref{fig:irregulartime} shows the performance change of different $n_{t'}$. Intuitively, when the $n_{t'}$ is extremely small, e.g. $n_{t'}=2$ and missing ratio equals $\frac{5}{6}$, the performance demonstrates a dramatical decrease, due to a massive loss of supervised labels.
In comparison, when the missing ratio is small, e.g. $n_{t'}=10$, the performance is comparable. Therefore, even if \textsc{STONet} has the ability to handle the temporally-irregular forecasting tasks, a large ratio of missing data in the temporal domain compromises the model performance.
\subsection{Further Ablation Study}\label{sec:ablation}
Several hyper-parameters affect the model performance, and in this part we try to explore their impacts.
First, we aim to figure out how the \textit{`embedding size'}  i.e. $d$, and the \textit{`layer number'} determine the expressivity of the model. 
Second, as demonstrated in \cite{li2020multipole}, the multi-leveled graphs in multipole graph neural network encoder help to increase the accuracy, because it enables each node to aggregate messages from farther nodes in spatial domains.
Thus, we attempt to figure out its effects.
Note that when the \textit{`number of levels'} of multipole graphs equals $1$, the graph kernel encoder in \textsc{STONet} is the same as proposed in \cite{simonovsky2017dynamic}.
Finally, we conduct analysis on $\alpha$, the \textit{`weight of the reconstruction loss'}. 
We give results on \texttt{Solar Energy} and \texttt{Temperature}, and the results on other datasets are shown in Appendix C.5.
\begin{figure}[htb]\vspace{-0.6cm}
    \centering
            \subfigure[Impacts of `\textit{embedding size}'.]{ \label{fig:ablationsolar}
                \includegraphics[height=0.322\columnwidth]{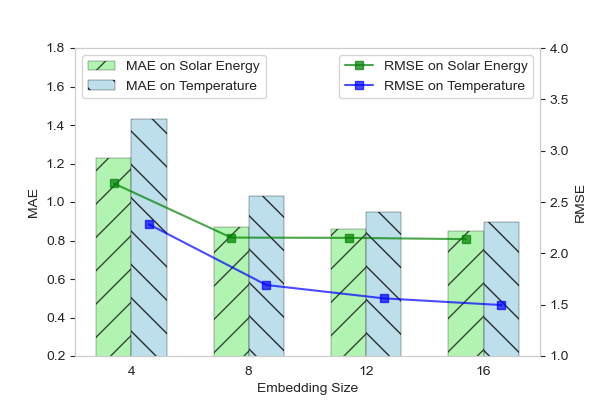}}\hspace{0mm}
            \subfigure[Impacts of `\textit{layer number}'.]{ \label{fig:ablationtemperature}
                \includegraphics[width=0.484\columnwidth]{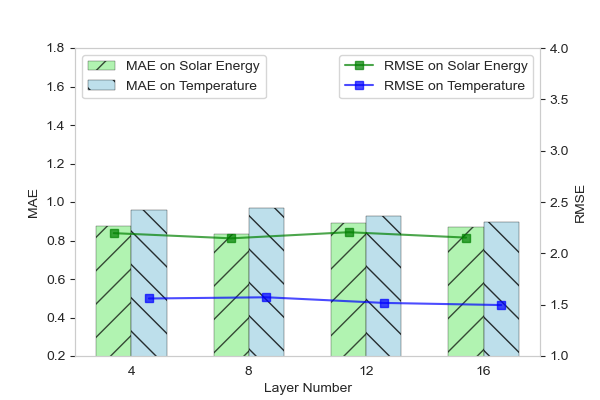}}\hspace{0mm} \vspace{-0.3cm}
    \caption{The change of MAE and RMSE on different hyper-parameters on Dataset of \texttt{Solar Energy} and \texttt{Temperature}. }\vspace{-0.4cm} \label{fig:ablation}
\end{figure}\vspace{-0.4cm}
    \begin{table}[htb]\caption{Comparison on different `\textit{number of levels}' and `\textit{loss weight}' on Dataset of \texttt{Solar Energy} and \texttt{Temperature}. } \label{table:ablation}
        \resizebox{1.00\columnwidth}{!}{
            \begin{tabular}{ccccc}
                \toprule
                \multicolumn{1}{c}{Datasets}           & \multicolumn{2}{c}{\texttt{Solar Energy}}                   & \multicolumn{2}{c}{\texttt{Temperature}}                    \\
                \midrule
                \textit{\# of levels} & \multicolumn{1}{c}{MAE} & \multicolumn{1}{c}{RMSE} & \multicolumn{1}{c}{MAE} & \multicolumn{1}{c}{RMSE} \\
                \cmidrule{1-1} \cmidrule(lr){2-3} \cmidrule(lr){4-5}
                1                                      & 1.0630{\scriptsize±0.0359}           & 2.5188{\scriptsize±0.0490}            & 0.8917{\scriptsize±0.0164}           & 1.4825{\scriptsize±0.0423}            \\
                2                                      & 0.9729{\scriptsize±0.2378}           & 2.3364{\scriptsize±0.2913}            & 0.8718{\scriptsize±0.0302}           & 1.4311{\scriptsize±0.0571}            \\
                3                                      & 0.8699{\scriptsize±0.0207}           & 2.1533{\scriptsize±0.0305}            & 0.8972{\scriptsize±0.0230}           & 1.4963{\scriptsize±0.0422}            \\
                \midrule
                
                \multicolumn{1}{c}{Datasets}   & \multicolumn{2}{c}{\texttt{Solar Energy}}                   & \multicolumn{2}{c}{\texttt{Temperature}}                    \\
                \midrule
                \textit{loss weight}       & \multicolumn{1}{c}{MAE} & \multicolumn{1}{c}{RMSE} & \multicolumn{1}{c}{MAE} & \multicolumn{1}{c}{RMSE} \\
                \cmidrule{1-1}   \cmidrule(lr){2-3} \cmidrule(lr){4-5}
                0.0                                    & 0.9195{\scriptsize±0.0102}           & 2.1677{\scriptsize±0.0209}            & 0.8845{\scriptsize±0.0196}           & 1.4837{\scriptsize±0.0313}            \\
                0.5                                    & 0.8699{\scriptsize±0.0399}           & 2.1533{\scriptsize±0.0563}            & 0.8972{\scriptsize±0.0230}           & 1.4963{\scriptsize±0.0422}            \\
                1.0                                    & 0.9159{\scriptsize±0.0430}           & 2.1724{\scriptsize±0.0625}            & 0.9692{\scriptsize±0.0216}           & 1.5880{\scriptsize±0.0392}            \\
                \bottomrule
            \end{tabular}
        }\vspace*{-0.3cm}
        \end{table}
        
\section{Conclusion}
An operator-driven spatio-temporal forecasting network is proposed, with its encoder based on graph kernel operator and decoder based on universal operator. Experiments show that it achieves improved performance in continuous physical quantities forecasting on spatial points of \textit{spatially-irregular distribution} , and allows both \textit{spatially-inductive} and \textit{temporally-irregular forecasting}.
\bibliography{STONet}
\bibliographystyle{icml2021}

\clearpage

\appendix
\section*{A. Notation and Method Supplementary}
\subsection*{A.1. Glossary of notations}
\begin{table}[ht]
    \centering
    \resizebox{1.05\columnwidth}{!}{
    \begin{tabular}{lp{0.73\columnwidth}}
    \toprule
    Symbol                                                                                                                                  & Used for      \\    
    \midrule
    $t$                                                                                                                                     & Time or timestamps.                                                                    \\
    $\bm{\mathrm{x}}$                                                                                                                            & Spatial locations.                                                                 \\
    $\bm{\mathrm{x}}_{\mathrm{new}}$                                                                                                                      & Spatial locations which are unseen by models during training.                                                                 \\
    $u$                                                                                                                                     & Physical quantities or continuous signals with spatial location and time as input .                    \\                                                         
    $N(\bm{\mathrm{x}})$                                                                                                                         & neighborhood of $\mathrm{x}$. \\
    $F$                                                                                                                                     & Mapping of historical observations into future predictions.                                                            \\
    ${a_t}$                                                                                                                                 &  Parametric function in  parametric PDEs.                                                  \\
    $\mathcal{L}_{a_t}$                                                                                                                     & Differential operator in  parametric PDEs.                                                  \\
    $\mathcal{F}$                                                                                                                           & Operator for solving the parametric PDEs, mapping paramteric function to solution function.                                                    \\
    $\mathcal{F^\dag}$                                                                                                                      & Operator for approximating true operators in the parametric PDEs.             \\                                       
    $\mathcal{K}_{a_t}$                                                                                                                      & Kernel operator as an instance of $\mathcal{F^{\dag}}$ for solving the  parametric PDEs.                                                    \\   
    $\mathcal{G}$                                                                                                                           & Operator of groundtruth.                                                  \\
    $\mathcal{G}^\dag$                                                                                                                      & Universal Operator to approximate the operator of groundtruth.                                                  \\
    $c,\xi,\theta,\eta$                                                                                                                     & Parameters in universal operator. \\
    $v^{\mathrm{enc}}$                                                                                                                                     & Solution representation function obtained by encoders.                              \\
    $v^{\mathrm{dec}}$                                                                                                                          & Solution representation function obtained by decoders.               \\
    $P$                                                                                                                              & Projector to map $u$ into $v$.               \\
    $P'$                                                                                                                              & Projector to map back $v$ into $u$.               \\                                                                                                                            
    \bottomrule
    \end{tabular}
    }
    \vspace*{0.2cm}
    \caption{Glossary of Notations used in this paper.}
    \end{table}
\subsection*{A.2. Multipole graph construction}
\paragraph{Neighborhood construction.}
The neighborhood system is constructed by $\epsilon$-ball, which can be written as for point $\bm{\mathrm{x}}$,
\begin{equation} 
    \begin{cases}
      &d(\bm{\mathrm{x}}, \bm{\mathrm{x}}_i) \leq \epsilon \quad \quad \bm{\mathrm{x}}_i\in N(\bm{\mathrm{x}});\\
      &d(\bm{\mathrm{x}}, \bm{\mathrm{x}}_i) > \epsilon \quad \quad \bm{\mathrm{x}}_i \not\in N(\bm{\mathrm{x}}).\\
    \end{cases}
\end{equation}
$d(\bm{\mathrm{x}}, \bm{\mathrm{y}})$ is the distance between point $\bm{\mathrm{x}}$ and $\bm{\mathrm{y}}$. In the planar dataset, the $d(\bm{\mathrm{x}}, \bm{\mathrm{y}})$ is calculated by 
\begin{align}
    d(\bm{\mathrm{x}}, \bm{\mathrm{y}}) = ||\bm{\mathrm{x}} - \bm{\mathrm{y}}||_2,
\end{align}
while in the sphere dataset, the distance term is calculated by
\begin{align}
    d(\bm{\mathrm{x}}, \bm{\mathrm{y}}) = \arccos(<\bm{\mathrm{x}} , \bm{\mathrm{y}}>).
\end{align}
\paragraph{Multipole graph kernel algorithm.}
The multipole method is to use a series of sparse and low-rank matrix to approximate the true kernel matrix. The true kernel matrix is decomposed into a hierarchy of low-rank structures.
First, all the spatial points are randomly divided into $L$ levels, and for each level $1\leq l \leq L$, the graph message passing is firstly operated inter-level, as
\begin{align}
    v_{t}^{\mathrm{enc},(l)}(\bm{\mathrm{x}}) = (\mathcal{K}_{a_t}v^{(l)}_t)(\bm{\mathrm{x}}) ,  
\end{align} 
Then, the low-level points' representation will be used to update the high-level one, as
\begin{align}
    v_{t}^{\mathrm{enc},(l+1)}(\bm{\mathrm{x}}) = (\mathcal{K}_{a_t}v^{(l)}_t)(\bm{\mathrm{x}}),   
\end{align} 
and the high-level points' representation will be also used to update the points of low level reversely, as
\begin{align}
    v_{t}^{\mathrm{enc},(l)}(\bm{\mathrm{x}}) = (\mathcal{K}_{a_t}v^{(l+1)}_t)(\bm{\mathrm{x}}).  
\end{align} 
It can be regarded as decomposing the kernel with a series of low-rank matrix by recursively applying the three steps.
And for each point, it can aggregate messages from neighbors of different levels, and thus the recursive structure of the algorithm can allow each point to be affected farther points, although it consumes more time.

\section*{B. A ODE-based Decoder}
We establish another decoder based on Neural ODE \cite{chen2019neural}. The ODE-based decoder is constructed by 
\begin{align}
    \frac{\partial u(\bm{\mathrm{x}},t)} {\partial t}  = (\mathcal{\bm{L}}_{a_t} u_t)(\bm{\mathrm{x}}). \\
\end{align}
In this way, we first use a universal operator $\mathcal{\bm{F}}$ to approximate $\mathcal{\bm{L}}_{a_t}$, and then use the Neural ODE, reads 
\begin{align}
    \frac{d u(\bm{\mathrm{x}},t)} {d t}  &= (\mathcal{\bm{F}} u_t)(\bm{\mathrm{x}}), \notag\\
     u(\bm{\mathrm{x}},t)  &= \int_0^t (\mathcal{\bm{F}} u_\tau)(\bm{\mathrm{x}})d\tau, \notag\\
\end{align}
For our model aims to handle the long-term prediction, the `\textit{backward}' process is extremely slow, thus we regard it as computational infeasible.
Table~\ref{tab:computecost} gives the comparison on ODE-based decoder and non-ODE-based decoder on the dataset of \texttt{Solar Energy}  and \texttt{Humidity} on one epoch, and the performance comparison.
Because the ODE-based decoder is extremely time-consuming, thus we did not conduct experiments on other datasets to show its effectiveness.
It shows that our model is superior to the ODE-based one in terms of both computational efficiency and prediction accuracy, because the over-fitting effects are extremely obvious in ODE-based one.
\begin{table}[htb]\caption{Comparison on different decoder type on Dataset of \texttt{Humidity} and \texttt{Solar Energy}. \textit{`Time'} is the one epoch training time cost by the method. The test is implemented on a single Nvidia-V100(32510MB).} \label{tab:computecost}
    \resizebox{1.00\columnwidth}{!}{
        \begin{tabular}{c|ccccc}
            \toprule
            \multicolumn{1}{c}{Datasets}           & \multicolumn{5}{c}{\texttt{Humidity}}                                      \\
            \midrule
            \textit{Decoder Type} & set&\multicolumn{1}{c}{MAE} & \multicolumn{1}{c}{RMSE} & \multicolumn{1}{c}{Memory} & \multicolumn{1}{c}{Time} \\
            \cmidrule{1-1} \cmidrule(lr){2-2} \cmidrule(lr){3-4} \cmidrule(lr){5-6}
            \multirow{2}{*}{\rotatebox{0}{ODE-based}}                     & train               & 3.1491{\scriptsize±0.4533}          & 7.1483{\scriptsize±0.0991}            & \multirow{2}{*}{13641MB}           & \multirow{2}{*}{$12^\prime37^{\prime \prime}$ }            \\
                                                                        & test                       &12.083{\scriptsize±6.2098}   & 17.2462{\scriptsize±9.5624} \\
            \midrule
            Non-ODE-based                 & test                 & 0.8699{\scriptsize±0.0399}           & 2.1533{\scriptsize±0.0563}            & 16163MB           &      $2^\prime19^{\prime \prime}$      \\
            \bottomrule

            \multicolumn{1}{c}{Datasets}           & \multicolumn{5}{c}{\texttt{Solar Energy}}                                      \\
            \midrule
            \textit{Decoder Type} & set&\multicolumn{1}{c}{MAE} & \multicolumn{1}{c}{RMSE} & \multicolumn{1}{c}{Memory} & \multicolumn{1}{c}{Time} \\

            \cmidrule{1-1} \cmidrule(lr){2-2} \cmidrule(lr){3-4} \cmidrule(lr){5-6}
            \multirow{2}{*}{\rotatebox{0}{ODE-based}}       & train                               & 0.6466{\scriptsize±0.0610}           & 0.8975{\scriptsize±0.1312}             & \multirow{2}{*}{22957MB}           & \multirow{2}{*}{$63^\prime07^{\prime \prime}$}            \\
                                                            & test                          &2.0103{\scriptsize±0.4933}   &2.1439{\scriptsize±0.3534}    \\
            \midrule
            Non-ODE-based                                   & test         & 0.4273{\scriptsize±0.0256}           & 0.6584{\scriptsize±0.0287}             & 25183MB           & $10^\prime42^{\prime \prime}$            \\
            \bottomrule
        \end{tabular}\vspace{-0.4cm}
    }
    \end{table}
\begin{figure*}[ht]
    \centering
            \subfigure[MPAE Comparision on Wave.]{ \label{fig:mapewave}
                \includegraphics[height=0.27\linewidth]{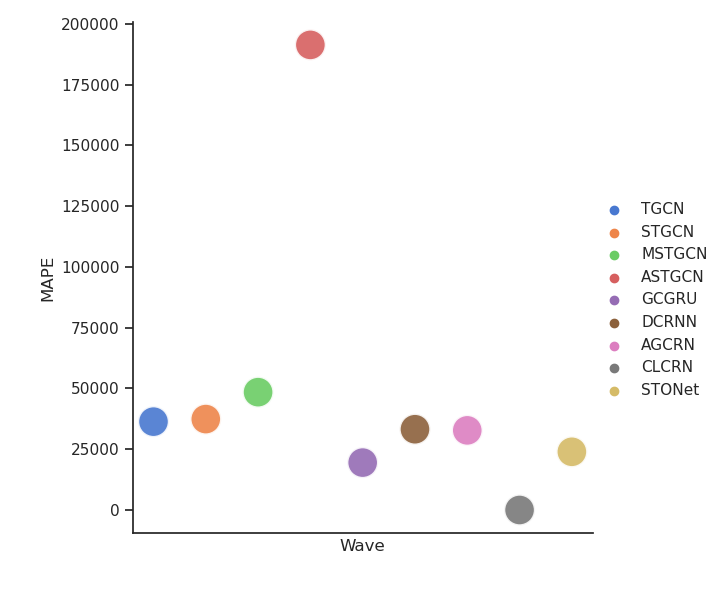}}\hspace{0mm}
            \subfigure[MAPE Comparison on Solar Energy.]{ \label{fig:mapesolar}
                \includegraphics[width=0.32\linewidth]{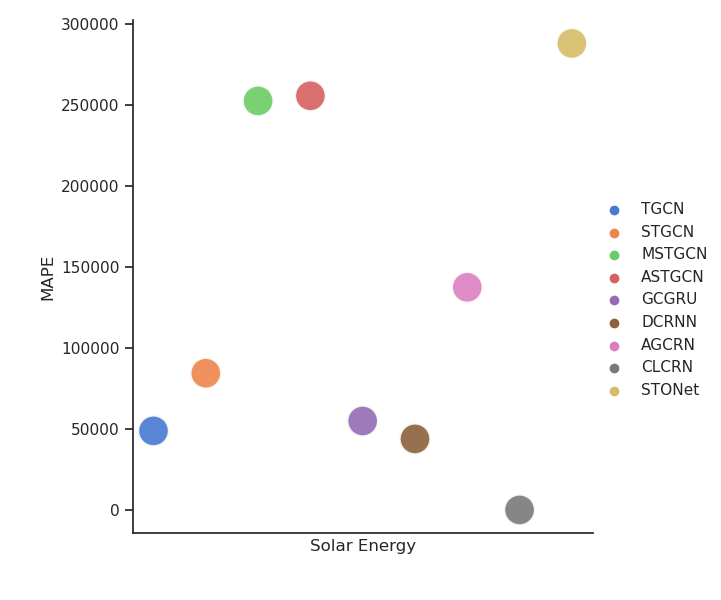}}\hspace{0mm}
            \subfigure[MAPE Comparison on Temperature.]{ \label{fig:mapetemperature}
                \includegraphics[width=0.32\linewidth]{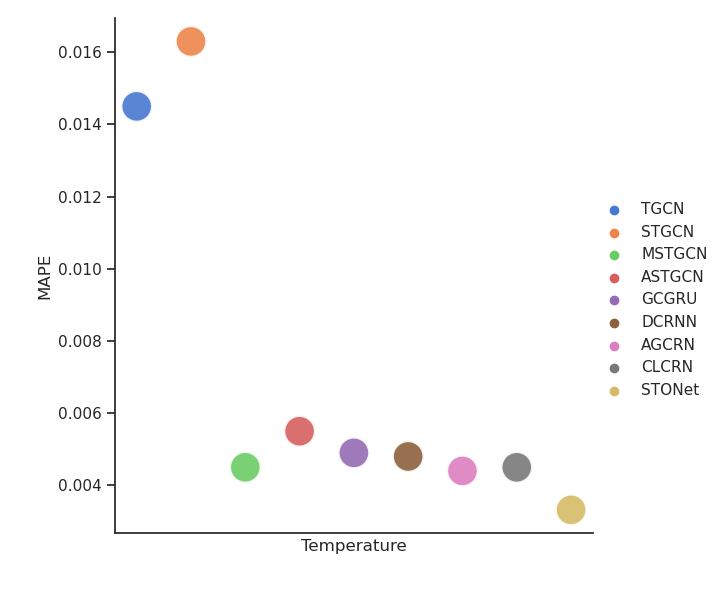}} \vspace{-0.3cm}\\
                \hspace{0.8mm}
            \subfigure[MPAE Comparision on Humidity.]{ \label{fig:mapehumid}
                \includegraphics[height=0.27\linewidth]{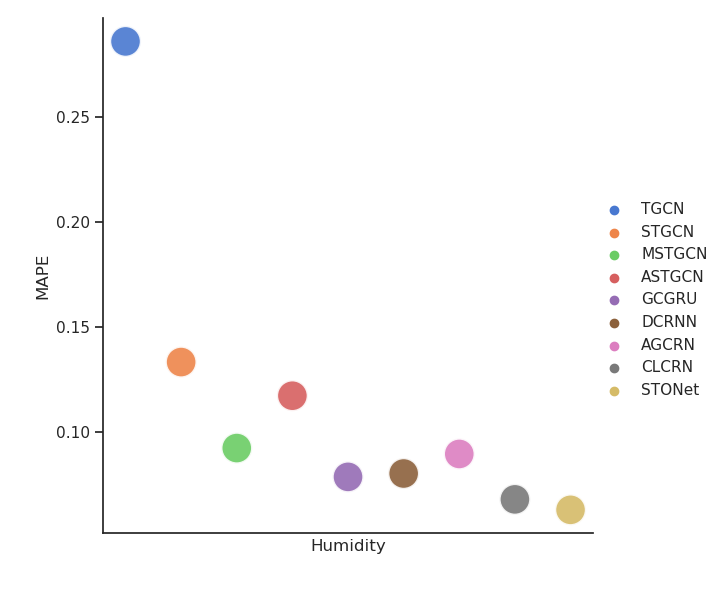}}\hspace{0mm}
            \subfigure[MAPE Comparison on Cloud Cover.]{ \label{fig:mapecloud}
                \includegraphics[width=0.32\linewidth]{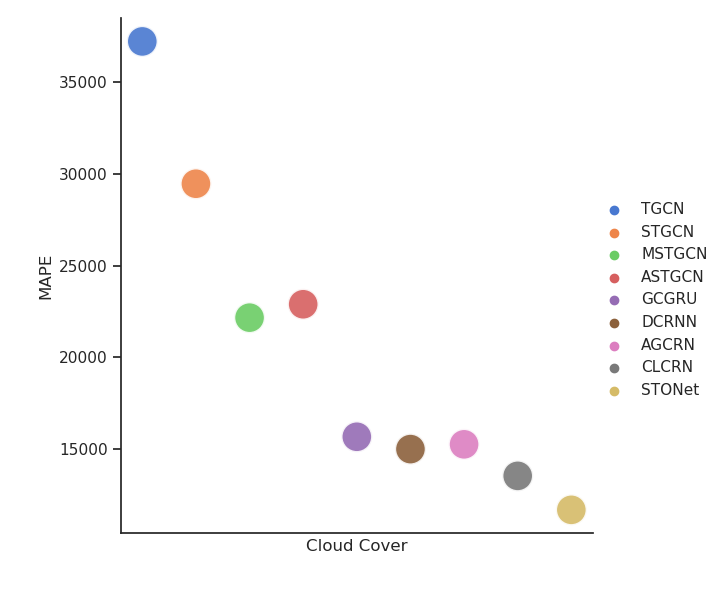}}\hspace{0mm}
            \subfigure[MAPE Comparison on Wind Component.]{ \label{fig:mapewind}
                \includegraphics[width=0.32\linewidth]{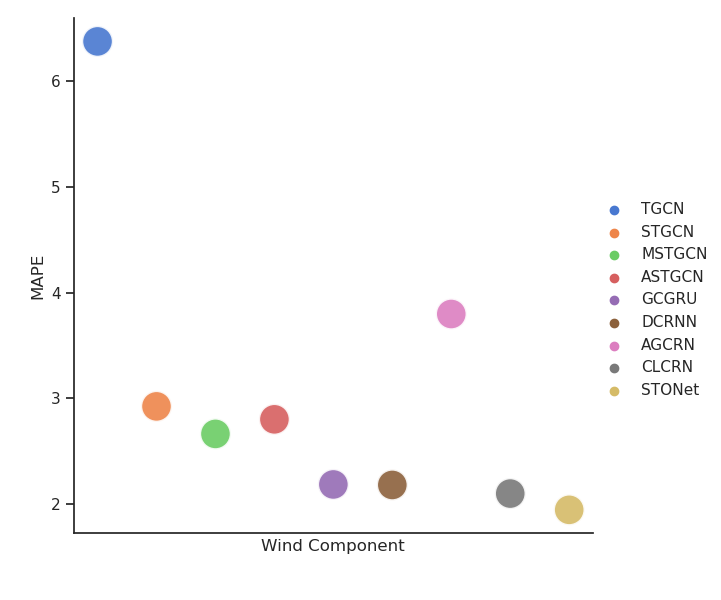}} \vspace{-0.3cm}
    \caption{Model Performance Comparison on MAPE.}\vspace{-0.4cm} \label{fig:mapecomparison}
\end{figure*}
\vspace{-0.4cm}
\section*{C. Experiment Supplementary}
\subsection*{C.1. Method description}
(1) \textsc{TGCN} is proposed for traffic forecasting, with the traffic sensors' graph constructed according to the road distance. The spatial convolution is based on Vanilla GCN, which uses the normed Laplacian matrix as the aggregation operator, and its temporal dynamics is modeled by GRU.\\
\\
(2) \textsc{STGCN} is also proposed for traffic forecasting. Different from \textsc{TGCN}, its temporal dynamics is modeled by 1DConv.\\
\\
(3) \textsc{MSTGCN} which is established for traffic forecasting, its spatial dependency is captured by ChebConv, which is a higher-order version of Vanilla GCN, approximating the graph spectral filters by Chebshev polynomials.\\
\\
(4) \textsc{ASTGCN} as a traffic forecasting model, is fully attention based, with spatial dependency captured by graph-attention mechanisms, and temporal dynamics captured by series self-attention.\\
\\
(5) \textsc{GCGRU} uses high order ChebConv to replace the linear transformation layers in GRU. It is established for many multi-variate time series forecasting tasks.\\
\\
(6) \textsc{AGCRN} trains the node embeddings, with the product of two node embeddings as their similarity, to adaptively construct the semantic graph as spatial dependency.  It is established for many multi-variate time series forecasting tasks, not only limited in traffic forecasting.\\
\\
(7) \textsc{CLCRN} as a spatio-temporal meteorological forecasting model, employs the conditional local convolution which is established for spherical datasets, and based on the assumption of smoothness of local patterns in weather forecasting. It is the state-of-the-art method for geophysical quantities forecasting.
\subsection*{C.2. Model comparision}
\paragraph{Metrics computation.}
Let 
\begin{align}
\mathbf{U}^{(t_1,t_{n_t})} = [u(\bm{\mathrm{X}},t_{1}), \ldots, u(\bm{\mathrm{X}},t_{n_t})]    \\
\mathbf{\hat U}^{(t_1,t_{n_t})} = [\hat u(\bm{\mathrm{X}},t_{1}), \ldots, \hat u(\bm{\mathrm{X}},t_{n_t}) ]
\end{align}
be the ground truth and the predictions obtained by neural networks respectively. The three metrics including MAE, RMSE and MAPE are calculated as

{\small
\begin{align*}
    \mathrm{MAE}(\mathbf{U}^{(t_1,t_{n_t})},\mathbf{\hat U}^{(t_1,t_{n_t})}) &= \frac{1}{n_t}\sum_{i=1}^{n_t} |\hat u(\bm{\mathrm{X}},t_{i}) - u(\bm{\mathrm{X}},t_{i}) |\\
    \mathrm{RMSE}(\mathbf{U}^{(t_1,t_{n_t})},\mathbf{\hat U}^{(t_1,t_{n_t})}) &= \frac{1}{n_t}\sqrt{\sum_{i=1}^{n_t} |\hat u(\bm{\mathrm{X}},t_{i}) - u(\bm{\mathrm{X}},t_{i}) |^2}\\
    \mathrm{MAPE}(\mathbf{U}^{(t_1,t_{n_t})},\mathbf{\hat U}^{(t_1,t_{n_t})}) &= \frac{1}{n_t}\sum_{i=1}^{n_t} \frac{|\hat u(\bm{\mathrm{X}},t_{i}) - u(\bm{\mathrm{X}},t_{i}) |}{|u(\bm{\mathrm{X}},t_{i}) |}\\
\end{align*}
}\vspace{-0.5cm}
\paragraph{MAPE comparison.}
Here we give the metrics of MAPE obtained by different methods, shown in Fig.~\ref{fig:mapecomparison}. MAPE metrics is not stable, because there exists a term in the denominator, and thus we do not consider the contributions of terms with ground truth equaling 0. However, for datasets \texttt{Wave}, \texttt{Solar Energy} and \texttt{Cloud Cover}, the minimal is still extremely small, causing the MAPE term extremely large. 
Therefore, we think the comparison of MAPE on these three datasets are not meaningful.

\subsection*{C.3. Spatially-inductive forecasting}
\begin{figure*}[ht]
    \subfigure[Change of metrics with \textit{`Ratio'} on Wave.]{ \label{fig:inductivewave}
        \includegraphics[width=1.0\linewidth]{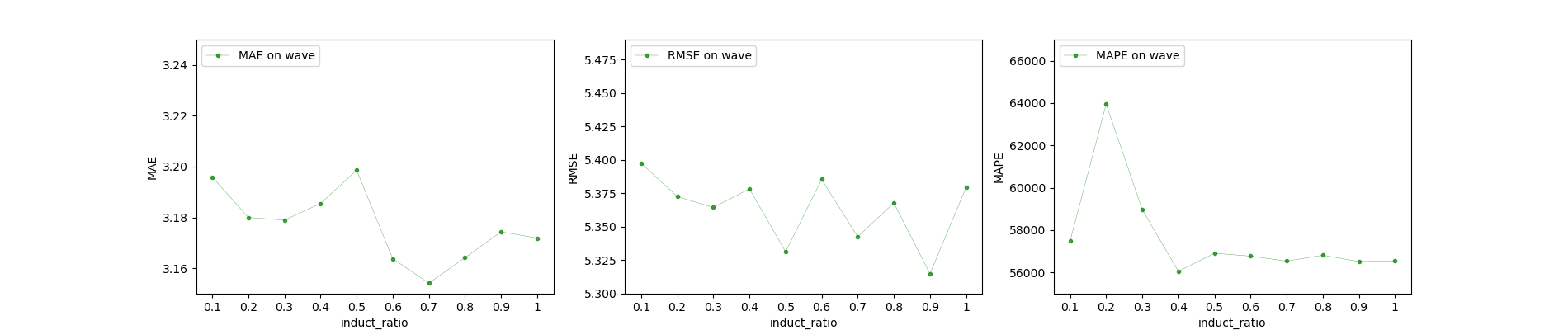}}\hspace{0mm}
    \subfigure[Change of metrics with \textit{`Ratio'} on Temperature.]{ \label{fig:inductivetemperature}
        \includegraphics[width=1.0\linewidth]{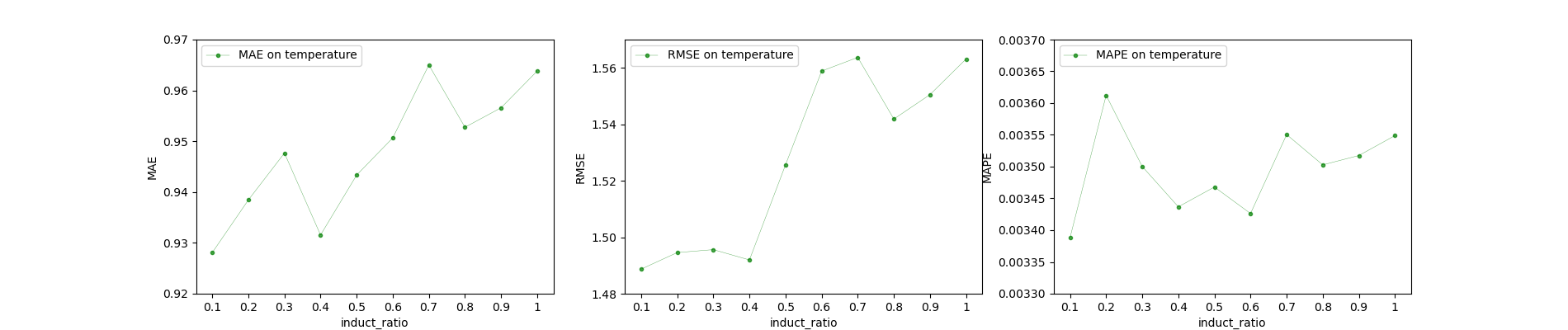}}\hspace{0mm}
    \subfigure[Change of metrics with \textit{`Ratio'} on Humidity.]{ \label{fig:inductivehumidity}
        \includegraphics[width=1.0\linewidth]{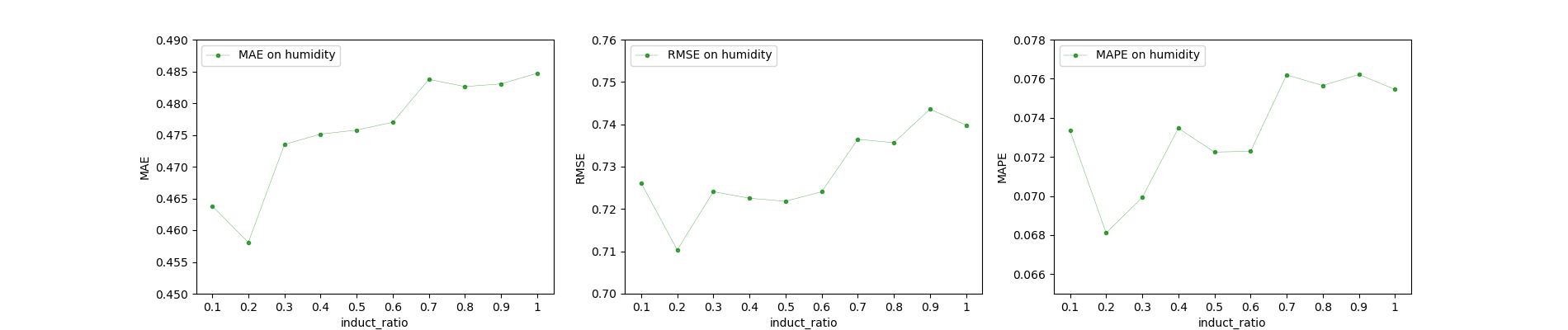}}\hspace{0mm}
    \subfigure[Change of metrics with \textit{`Ratio'} on Cloud Cover.]{ \label{fig:inductivecloud}
        \includegraphics[width=1.0\linewidth]{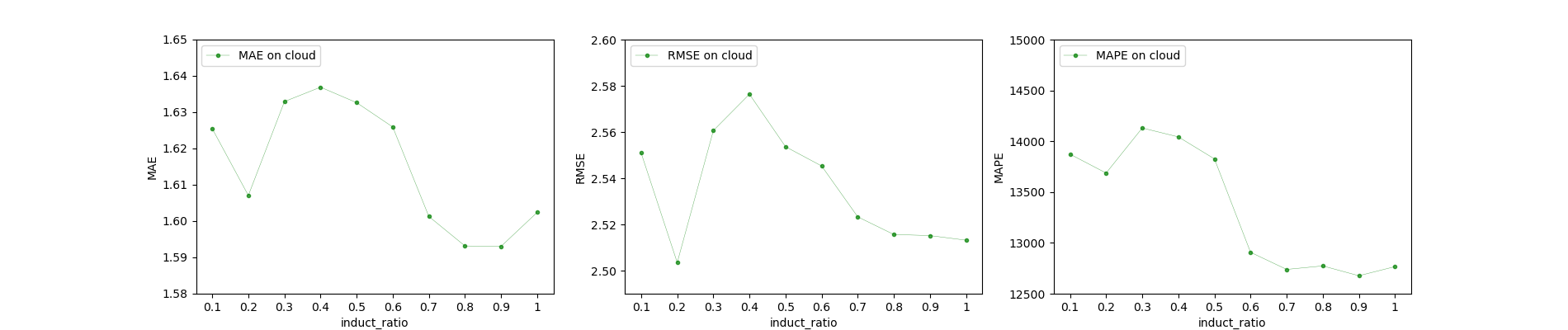}}\hspace{0mm}
    \subfigure[Change of metrics with \textit{`Ratio'} on Wind Component.]{ \label{fig:inductivewind}
        \includegraphics[width=1.0\linewidth]{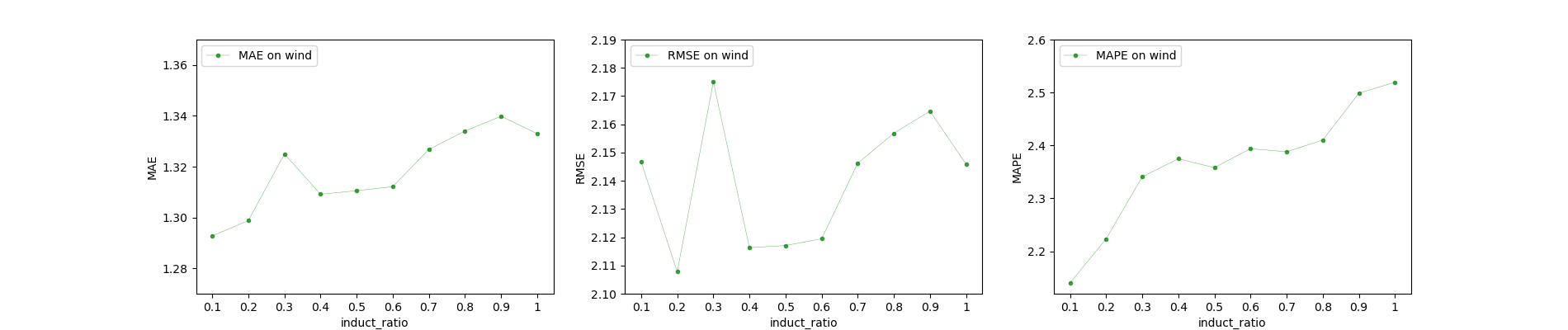}}\hspace{0mm}
    \caption{Model performance with the change of inductive points' \textit{`Ratio'}.}\vspace{-0.4cm} \label{fig:inductiveall}
\end{figure*}

\paragraph{Data preprocess.}
The nodes in each dataset need to be re-divided into nodes for training and nodes for inductive evaluation. The detailed processing is 
\begin{itemize}
    \item For \texttt{Wave}, The nodes are originally sampled from $64 \times 64$ images, so we randomly choose another $512$ nodes to evaluate models' spatially-inductive forecasting ability. 
    \item For \texttt{Solar Energy}: Since it is a dataset with low-resolution, we randomly choose $108$ nodes for training, and the rest $29$ nodes are used for inductive evaluation.
    \item Four the weather datasets: We first downsample the resolution to $16\times32$, with $512$ nodes used for training, and randomly choose another $512$ different nodes for inductive evaluation.       
\end{itemize}
Fig.~\ref{fig:inductiveall} gives further details on spatially-inductive forecasting on all datasets except \texttt{Solar Energy}, because the inductive ratio is always fixed for such a dataset of low spatial resolution.
It shows that with the increase of the inductive \textit{`Ratio'}, the prediction accuracy decreases. However, all the deviations are very small and acceptable.   
\subsection*{C.4. Temporally-irregular forecasting}
\begin{figure*}[ht]\centering
    \subfigure[Change of metrics with \textit{`Number of Output Timestamps'} on Wave.]{ \label{fig:temporalwave}
        \includegraphics[width=0.9\linewidth]{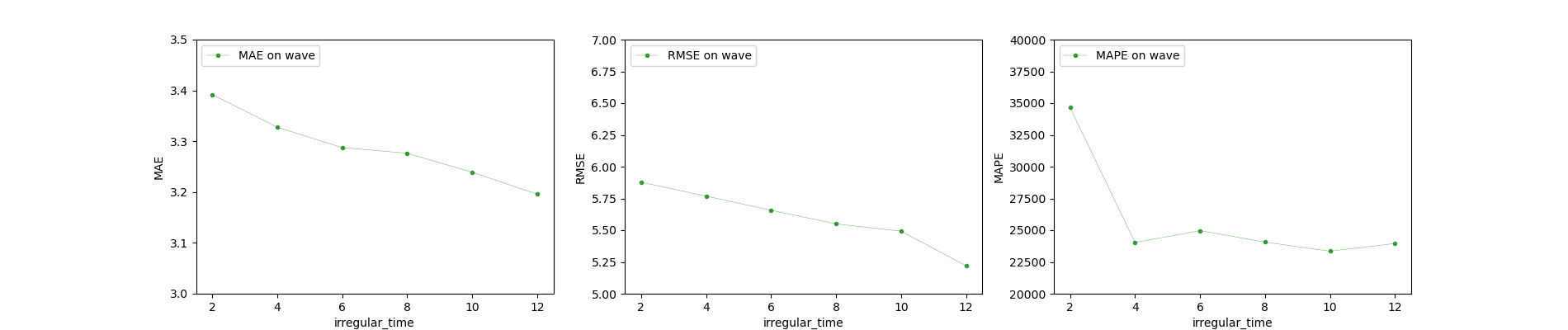}}\vspace{-0.35cm}
    \subfigure[Change of metrics with \textit{`Number of Output Timestamps'} on Solar Energy.]{ \label{fig:temporalsolar}
        \includegraphics[width=0.9\linewidth]{ablation_irregular_time_wave.png}}\vspace{-0.35cm}
    \subfigure[Change of metrics with \textit{`Number of Output Timestamps'} on Temperature.]{ \label{fig:temporaltemperature}
        \includegraphics[width=0.9\linewidth]{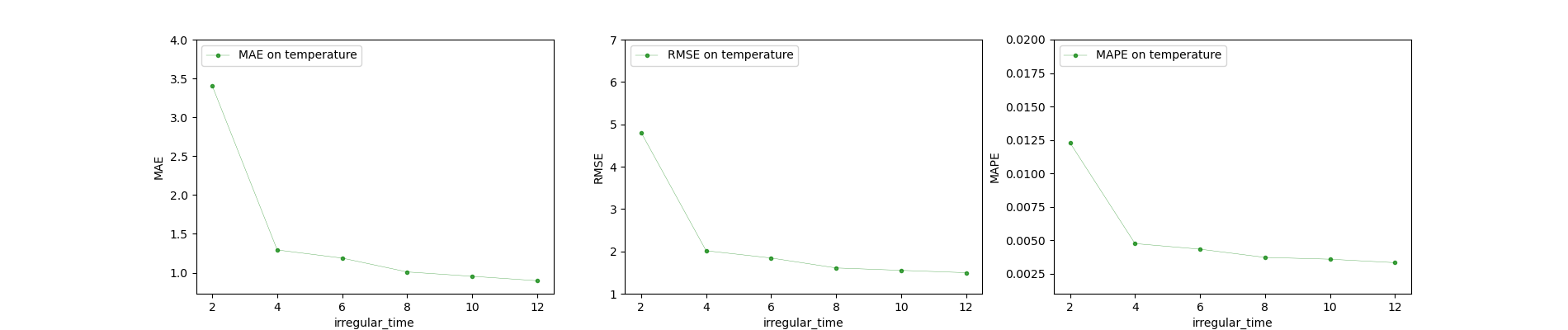}}\vspace{-0.35cm}
    \subfigure[Change of metrics with \textit{`Number of Output Timestamps'} on Humidity.]{ \label{fig:temporalhumidity}
        \includegraphics[width=0.9\linewidth]{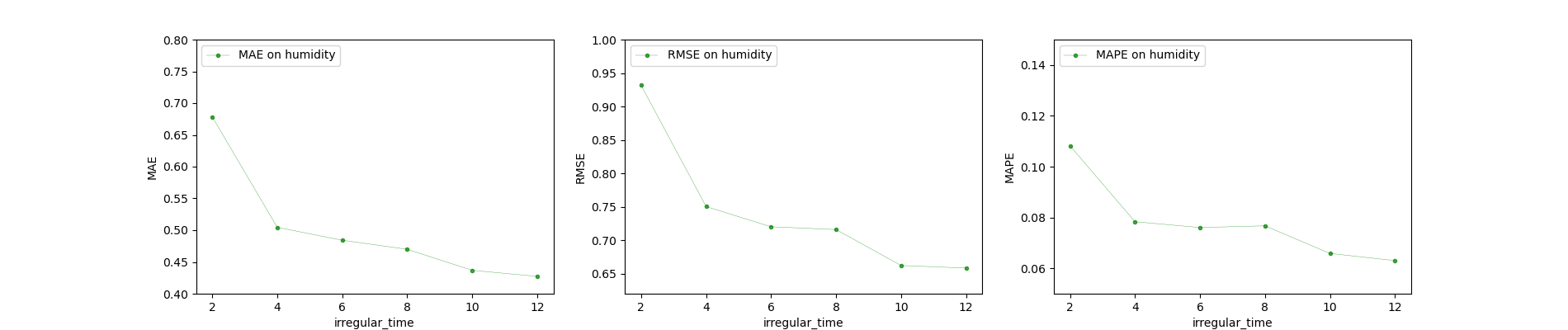}}\vspace{-0.35cm}
    \subfigure[Change of metrics with \textit{`Number of Output Timestamps'} on Cloud Cover.]{ \label{fig:temporalcloud}
        \includegraphics[width=0.9\linewidth]{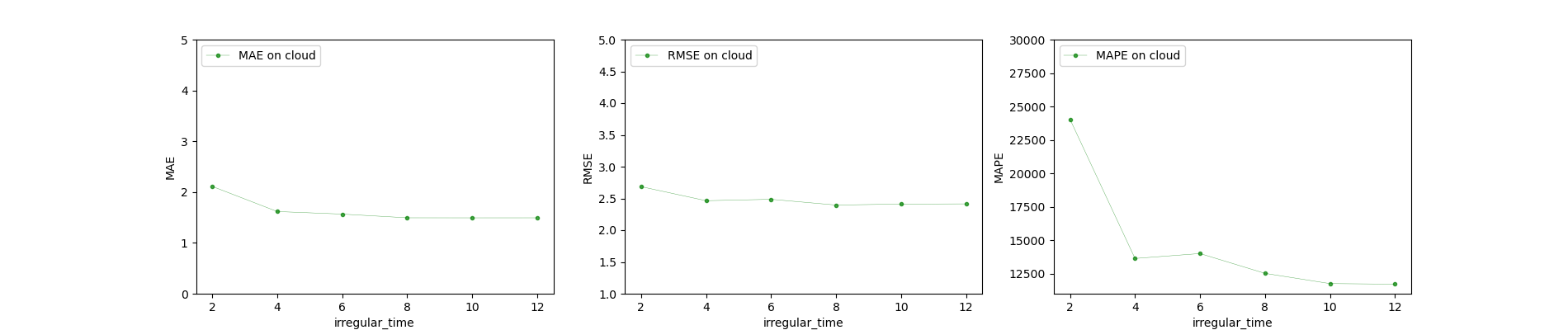}}\vspace{-0.35cm}
    \subfigure[Change of metrics with \textit{`Number of Output Timestamps'} on Wind Component.]{ \label{fig:temporalwind}
        \includegraphics[width=0.9\linewidth]{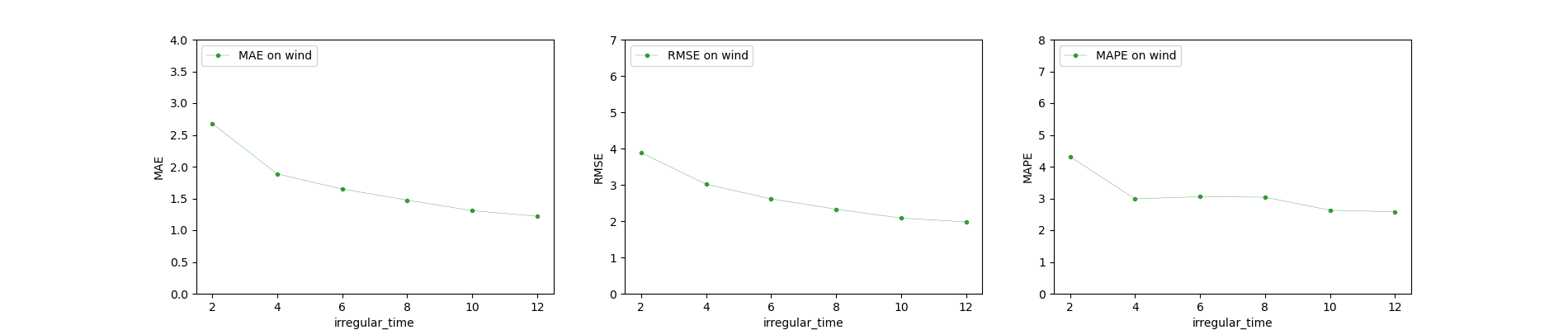}}\vspace{-0.35cm}
    \caption{Model performance with the change of \textit{`Number of Output Timestamps'}.}\vspace{-0.4cm} \label{fig:temporalall}
\end{figure*}

Fig.~\ref{fig:temporalall} gives further details on temporally-irregular forecasting on all datasets except \texttt{Solar Energy}.
As the output timestamps' labels fed in the model decrease, the predictive performance decrease due to excessive loss of supervised signals and information.
\subsection*{C.5. Ablation Study}
\begin{figure}[ht]\centering
    \subfigure[Change of metrics with \textit{`Embeding Size'} on Wave.]{ \label{fig:embedwave}
        \includegraphics[width=1.05\columnwidth]{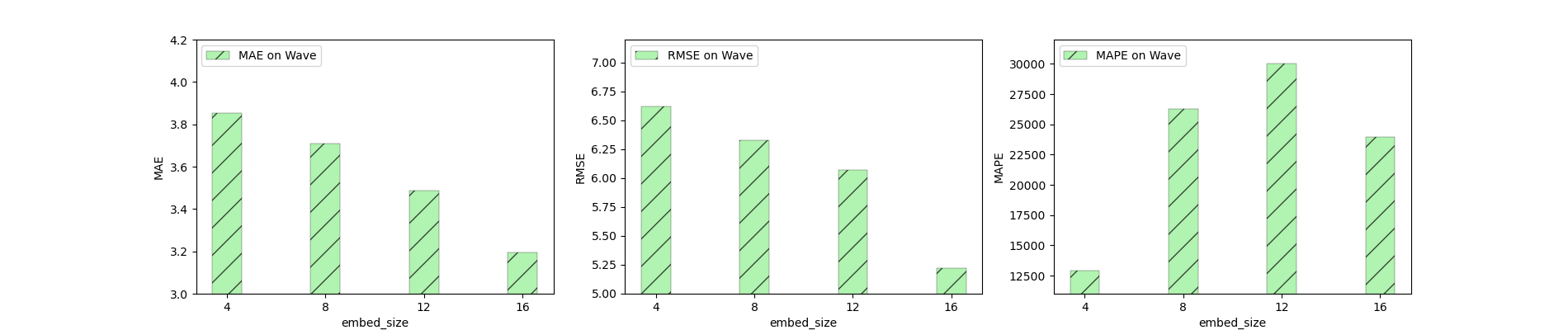}}\vspace{-0.4cm}
    \subfigure[Change of metrics with \textit{`Embeding Size'} on Solar Energy.]{ \label{fig:embedsolar}
        \includegraphics[width=1.05\columnwidth]{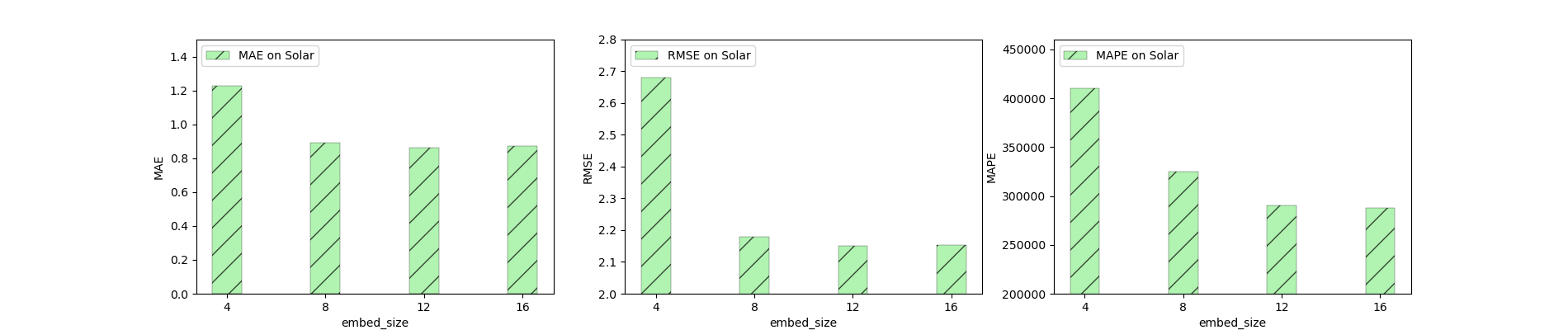}}\vspace{-0.4cm}
    \subfigure[Change of metrics with \textit{`Embeding Size'} on Temperature.]{ \label{fig:embedtemperature}
        \includegraphics[width=1.05\columnwidth]{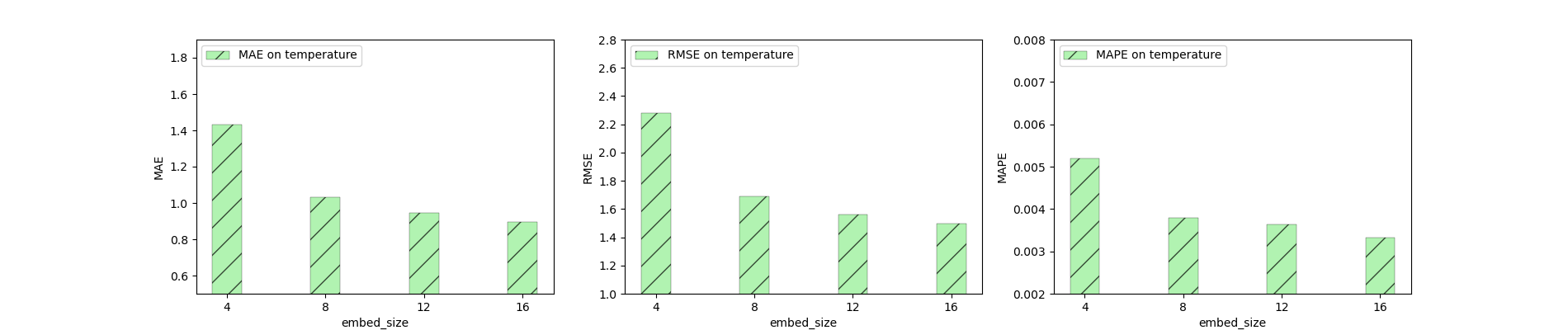}}\vspace{-0.4cm}
    \subfigure[Change of metrics with \textit{`Embeding Size'} on Humidity.]{ \label{fig:embedhumidity}
        \includegraphics[width=1.05\columnwidth]{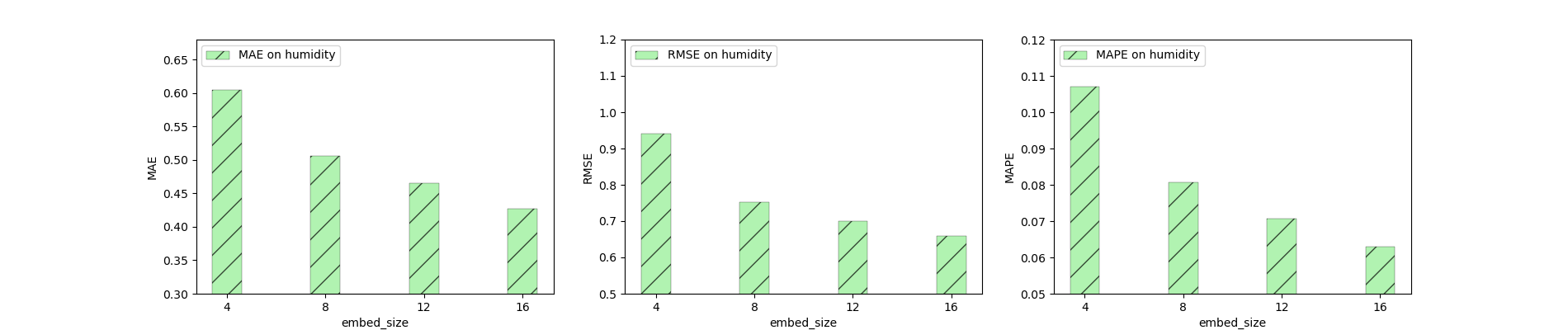}}\vspace{-0.4cm}
    \subfigure[Change of metrics with \textit{`Embeding Size'} on Cloud Cover.]{ \label{fig:embedcloud}
        \includegraphics[width=1.05\columnwidth]{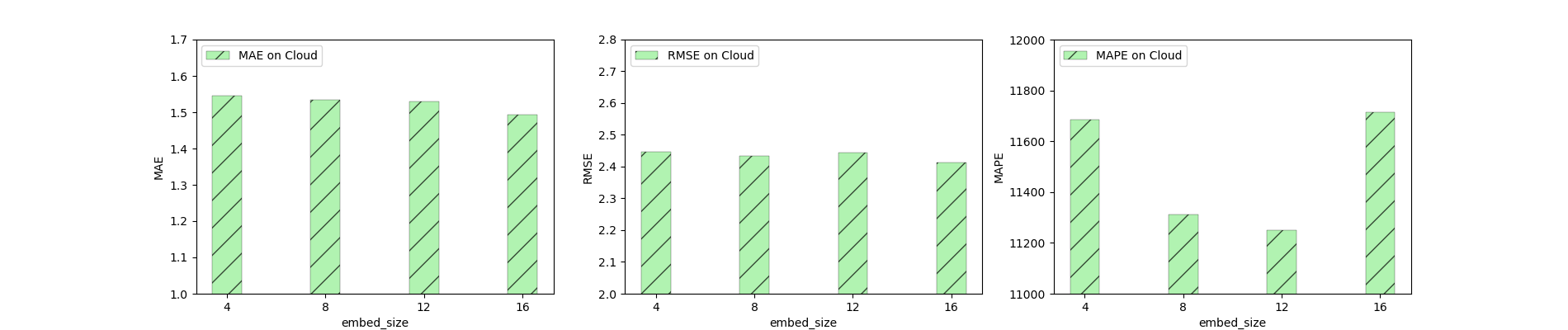}}\vspace{-0.4cm}
    \subfigure[Change of metrics with \textit{`Embeding Size'} on Wind Component.]{ \label{fig:embedwind}
        \includegraphics[width=1.05\columnwidth]{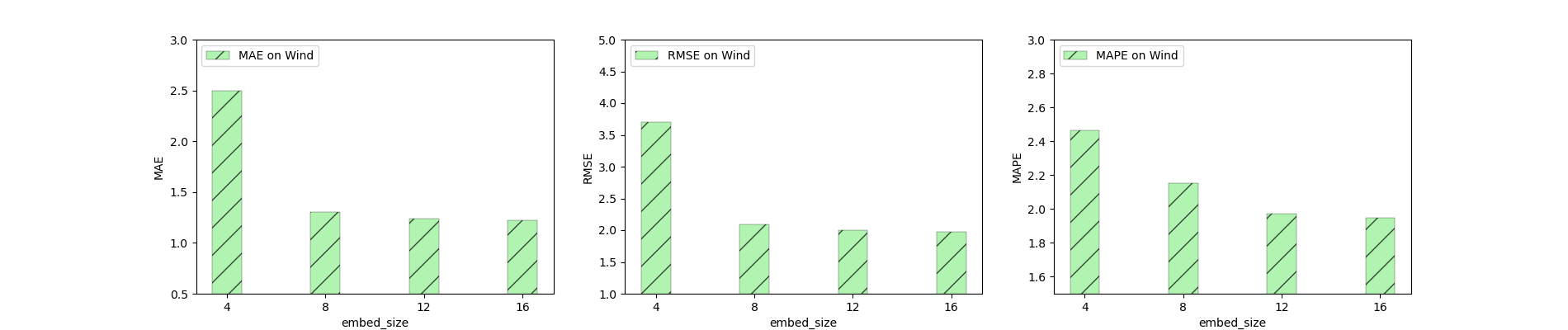}}\vspace{-0.4cm}
        \caption{Model Performance Comparison on \textit{`Embeding Size'}.}\vspace{-0.4cm} \label{fig:embedall}
\end{figure}

\begin{figure}[ht]\centering
    \subfigure[Change of metrics with \textit{`Layer Number'} on Wave.]{ \label{fig:layerwave}
        \includegraphics[width=1.05\columnwidth]{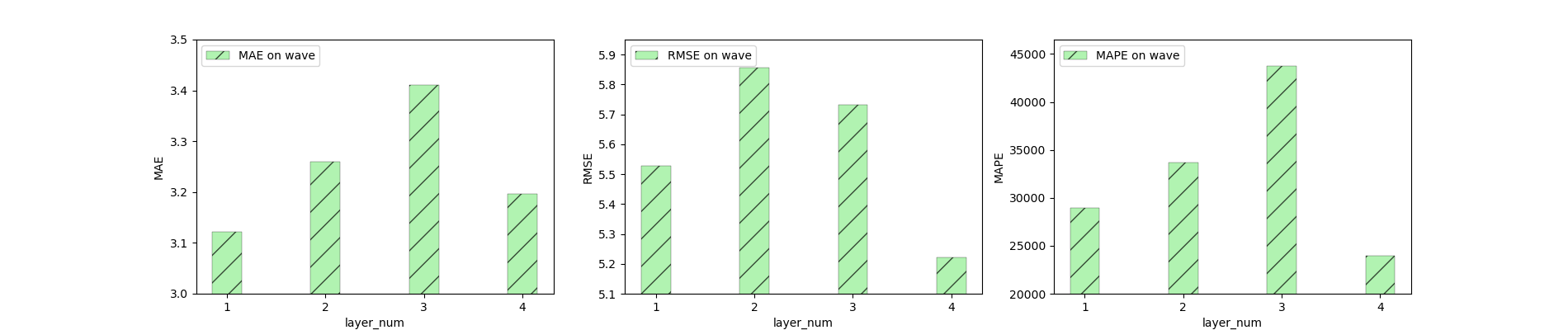}}\vspace{-0.3cm}
    \subfigure[Change of metrics with \textit{`Layer Number'} on Solar Energy.]{ \label{fig:layersolar}
        \includegraphics[width=1.05\columnwidth]{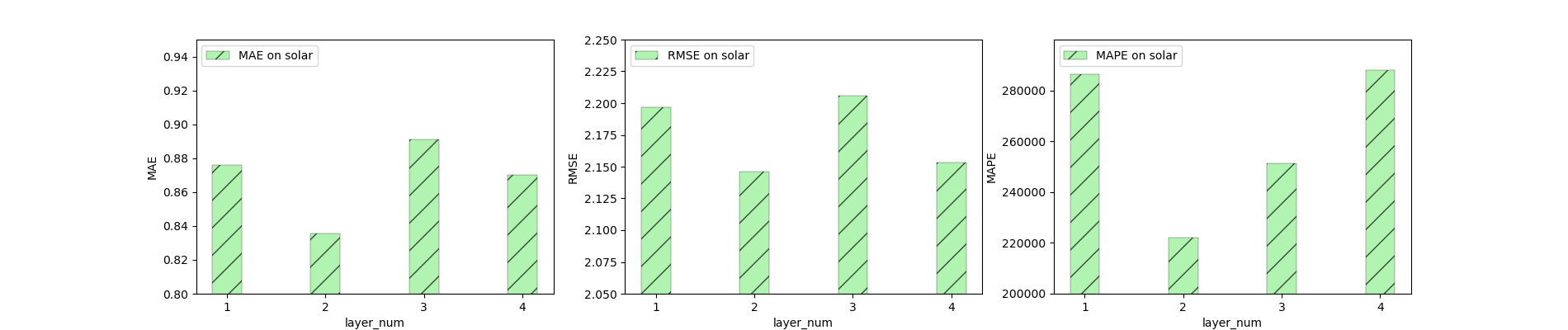}}\vspace{-0.3cm}
    \subfigure[Change of metrics with \textit{`Layer Number'} on Temperature.]{ \label{fig:layertemperature}
        \includegraphics[width=1.05\columnwidth]{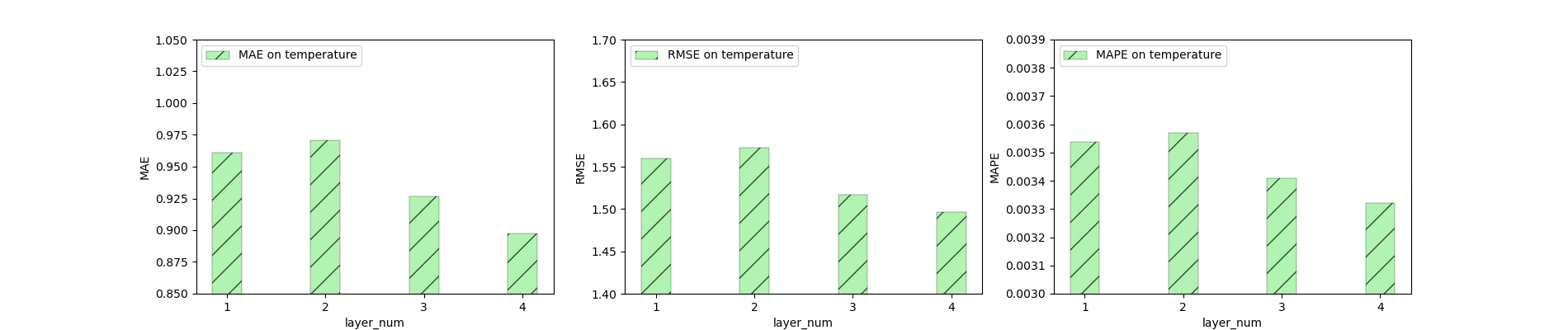}}\vspace{-0.3cm}
    \subfigure[Change of metrics with \textit{`Layer Number'} on Humidity.]{ \label{fig:layerhumidity}
        \includegraphics[width=1.05\columnwidth]{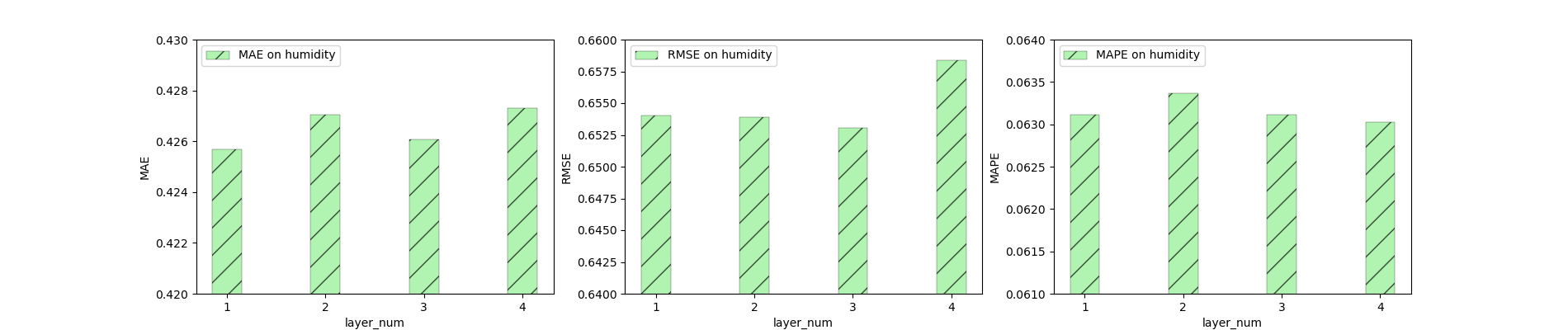}}\vspace{-0.3cm}
    \subfigure[Change of metrics with \textit{`Layer Number'} on Cloud Cover.]{ \label{fig:layercloud}
        \includegraphics[width=1.05\columnwidth]{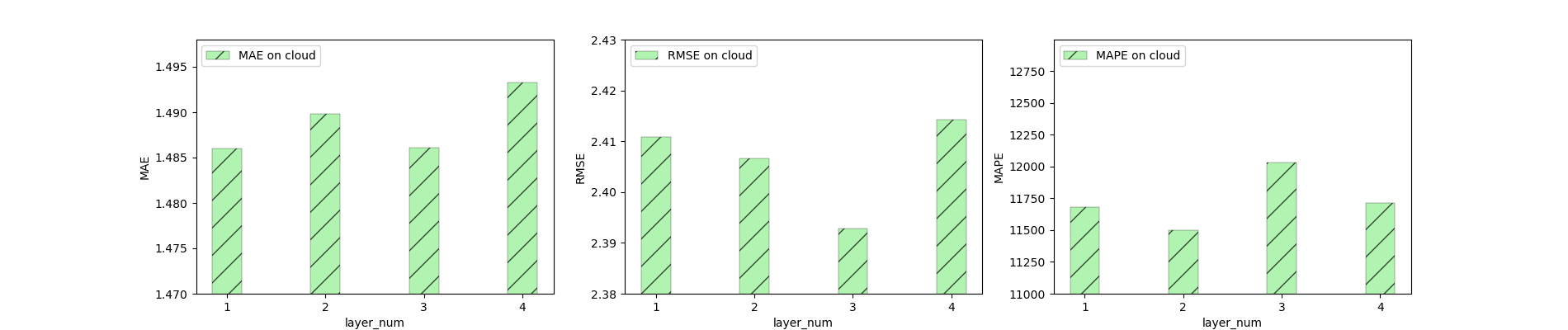}}\vspace{-0.3cm}
    \subfigure[Change of metrics with \textit{`Layer Number'} on Wind Component.]{ \label{fig:layerwind}
        \includegraphics[width=1.05\columnwidth]{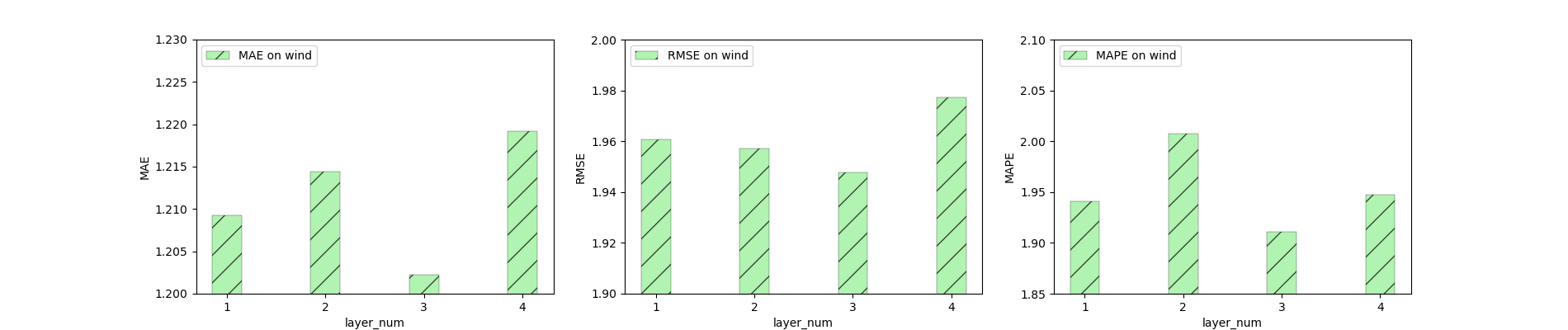}}\vspace{-0.3cm}
        \caption{Model Performance Comparison on \textit{`Layer Number'}.}\vspace{-0.3cm} \label{fig:layerall}
\end{figure}

\begin{figure}[ht]\centering
\subfigure[Change of metrics with \textit{`Level Number'} on Humidity.]{ \label{fig:levelhumidity}
\includegraphics[width=1.05\columnwidth]{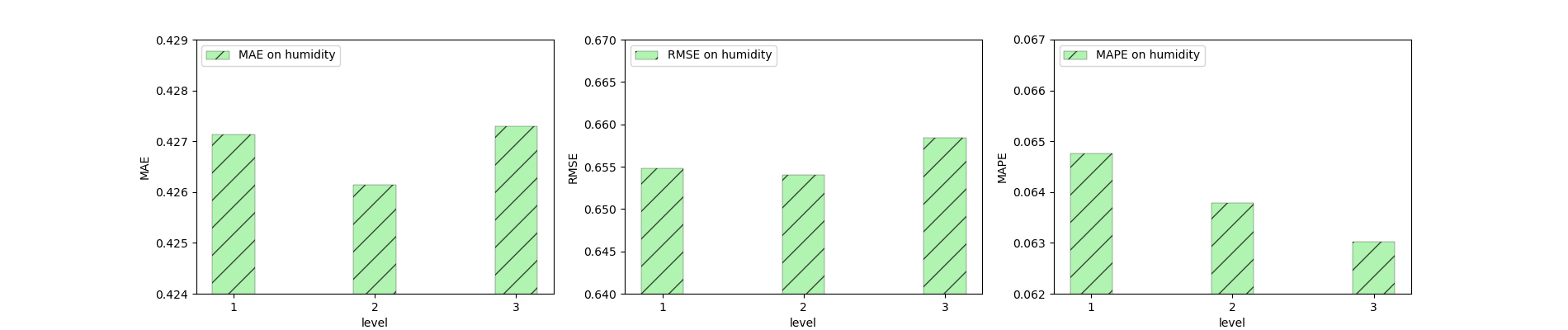}}\vspace{-0.3cm}
\subfigure[Change of metrics with \textit{`Level Number'} on Cloud Cover.]{ \label{fig:levelcloud}
\includegraphics[width=1.05\columnwidth]{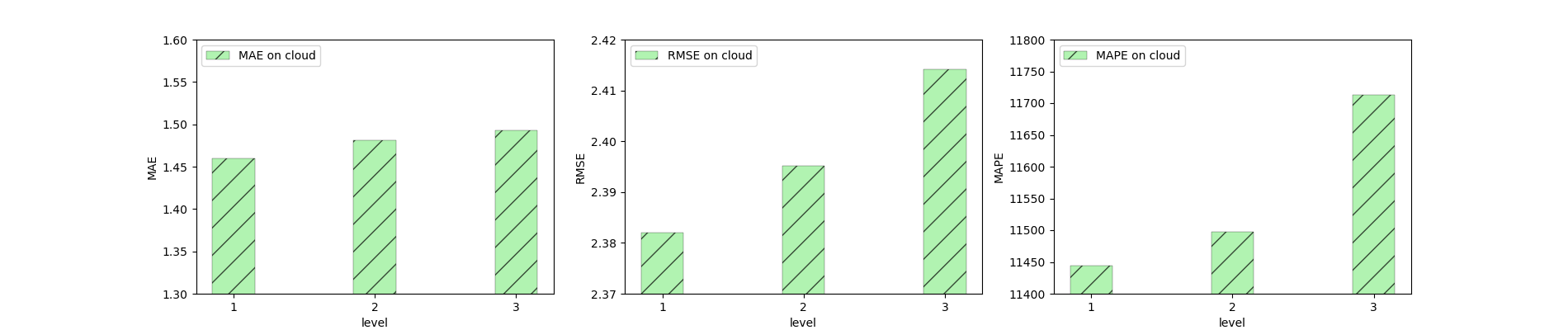}}\vspace{-0.3cm}
\subfigure[Change of metrics with \textit{`Level Number'} on Wind Component.]{ \label{fig:levelwind}
\includegraphics[width=1.05\columnwidth]{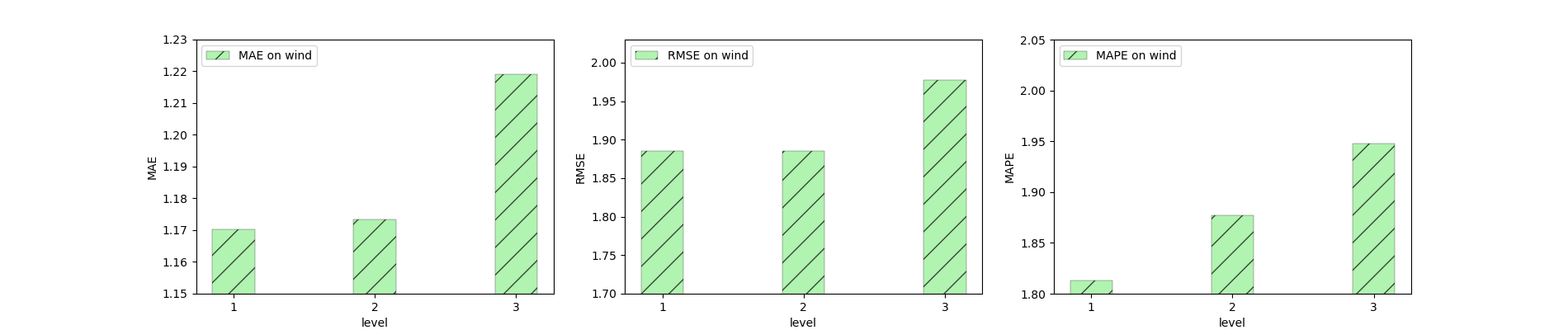}}\vspace{-0.3cm}
\caption{Model Performance Comparison on \textit{`Level Number'}.}\vspace{-0.3cm} \label{fig:levelall}
\end{figure}

\begin{figure}[ht]\centering
    \subfigure[Change of metrics with \textit{`Loss Weight'} on Wave.]{ \label{fig:weightwave}
        \includegraphics[width=1.05\columnwidth]{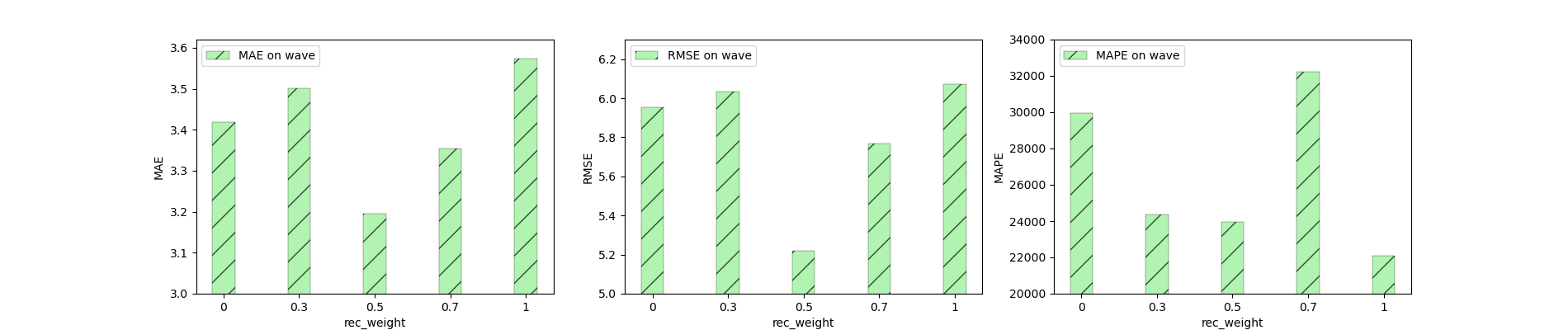}}\vspace{-0.3cm}
    \subfigure[Change of metrics with \textit{`Loss Weight'} on Solar Energy.]{ \label{fig:weightsolar}
        \includegraphics[width=1.05\columnwidth]{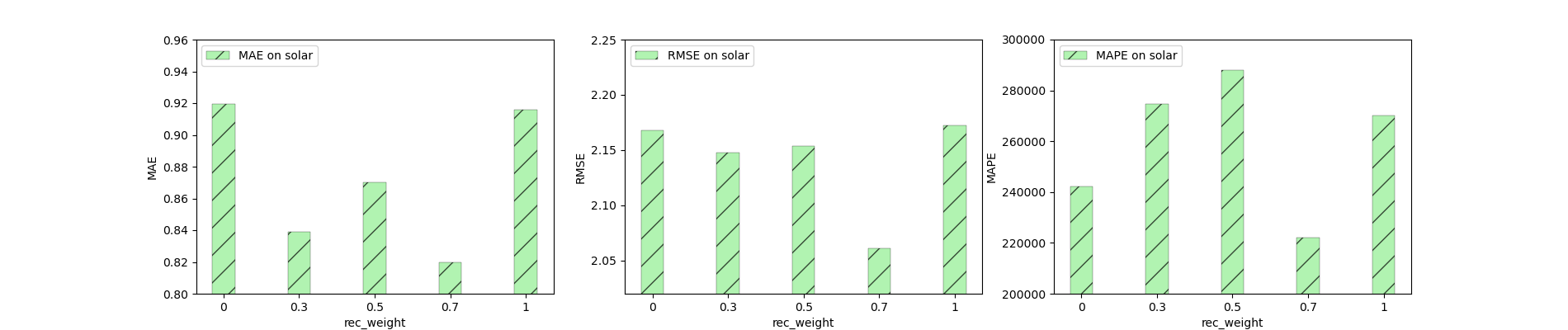}}\vspace{-0.3cm}
    \subfigure[Change of metrics with \textit{`Loss Weight'} on Temperature.]{ \label{fig:weighttemperature}
        \includegraphics[width=1.05\columnwidth]{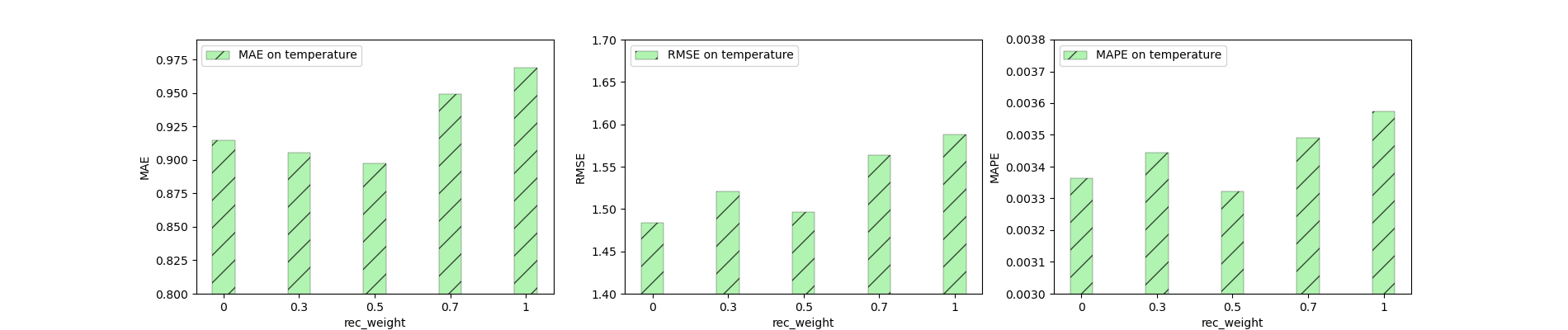}}\vspace{-0.3cm}
    \subfigure[Change of metrics with \textit{`Loss Weight'} on Humidity.]{ \label{fig:weighthumidity}
        \includegraphics[width=1.05\columnwidth]{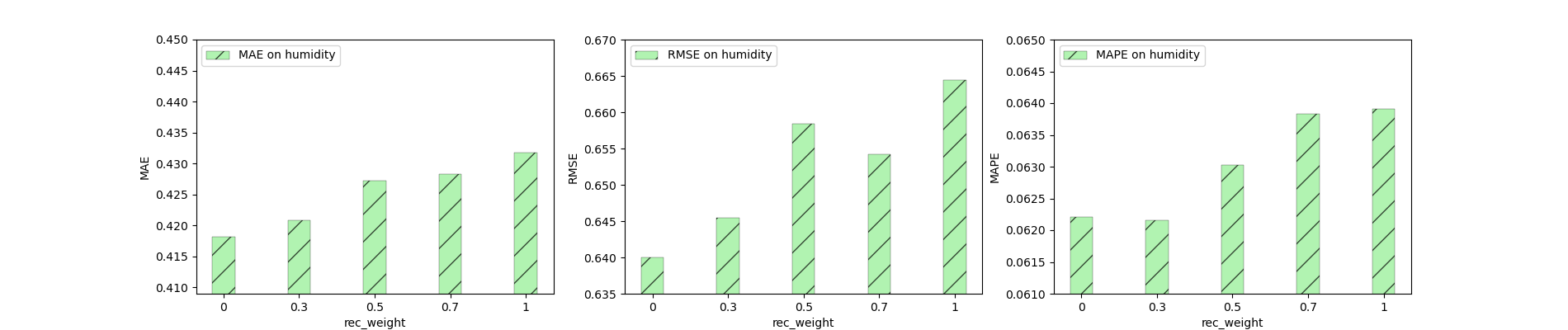}}\vspace{-0.3cm}
    \subfigure[Change of metrics with \textit{`Loss Weight'} on Cloud Cover.]{ \label{fig:weightcloud}
        \includegraphics[width=1.05\columnwidth]{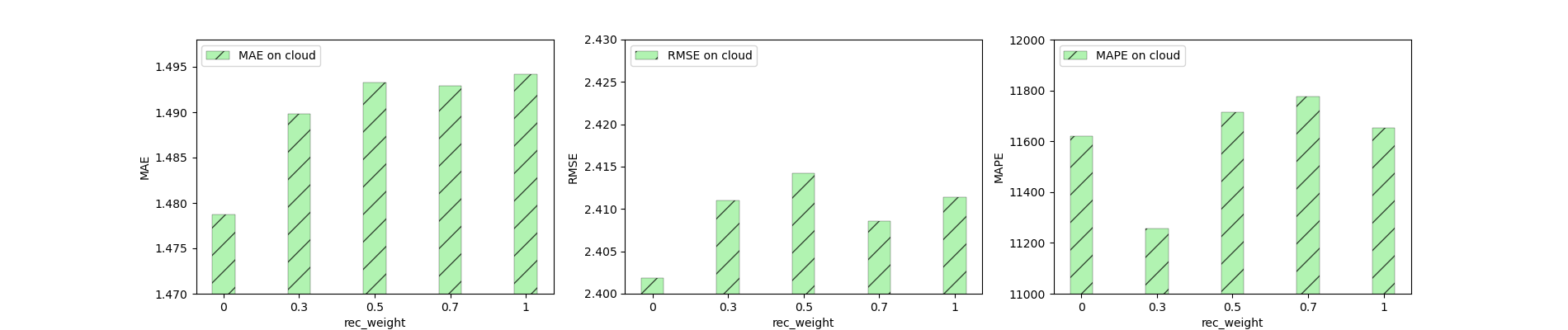}}\vspace{-0.3cm}
    \subfigure[Change of metrics with \textit{`Loss Weight'} on Wind Component.]{ \label{fig:weightwind}
        \includegraphics[width=1.05\columnwidth]{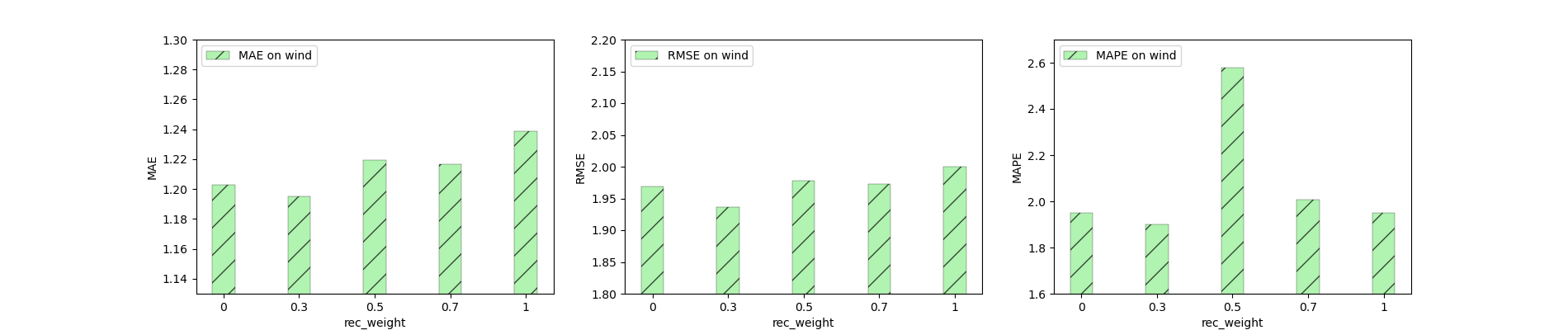}}\vspace{-0.3cm}
        \caption{Model Performance Comparison on \textit{`Loss Weight'}.}\vspace{-0.3cm} \label{fig:weightall}
\end{figure}

For each dataset, we give ablation study on the four hyper-parameters, which are \textit{`embedding size'}, \textit{`layer number'}, \textit{`level numbers'} and \textit{`weight of reconstruction loss'}.
The change of performance with the change of hyper-parameters is shown in Fig.~\ref{fig:embedall}, ~\ref{fig:layerall}, ~\ref{fig:levelall} and ~\ref{fig:weightall} respectively.
\vspace{-0.3cm}

\begin{figure}[H]\centering
    \subfigure[Change of metrics with \textit{`Level Number'} on Wave.]{ \label{fig:levelwave}
        \includegraphics[width=1.05\columnwidth]{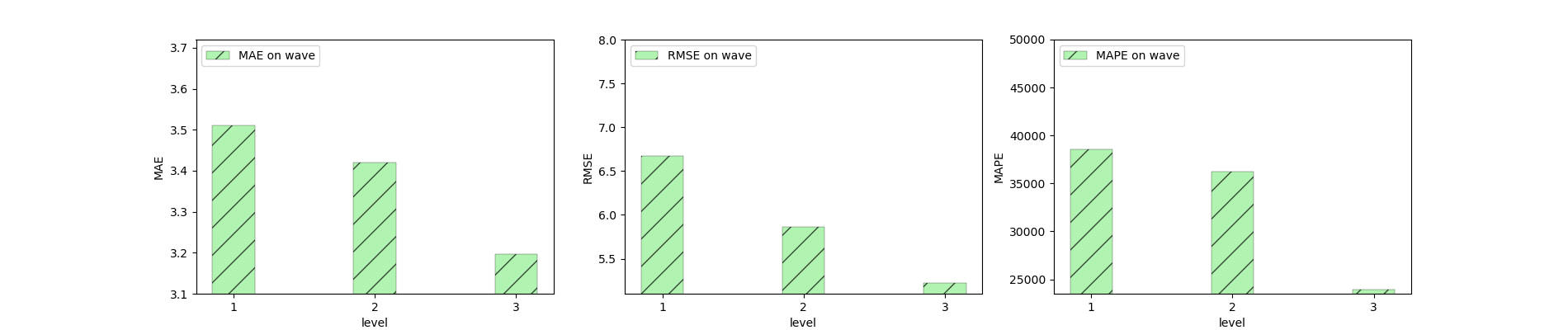}}\vspace{-0.3cm}
    \subfigure[Change of metrics with \textit{`Level Number'} on Solar Energy.]{ \label{fig:levelsolar}
        \includegraphics[width=1.05\columnwidth]{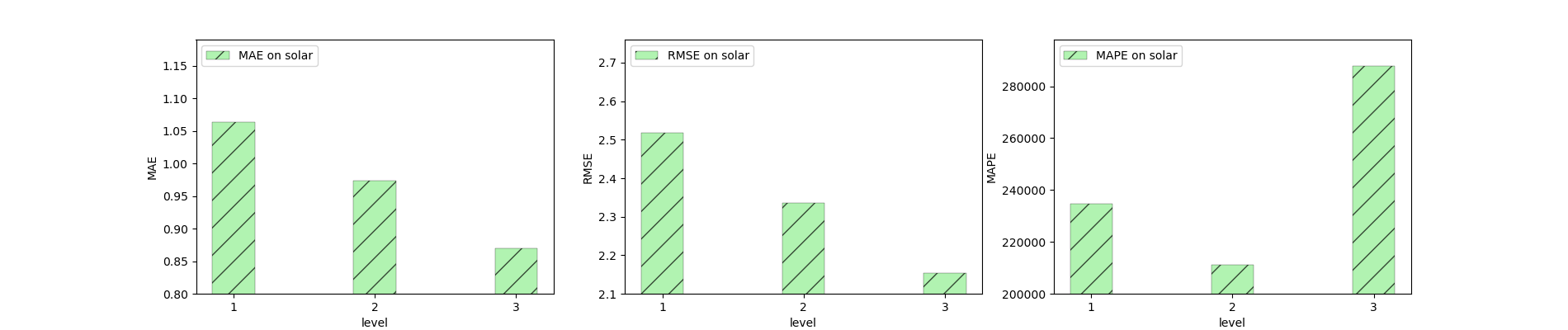}}\vspace{-0.3cm}
    \subfigure[Change of metrics with \textit{`Level Number'} on Temperature.]{ \label{fig:leveltemperature}
        \includegraphics[width=1.05\columnwidth]{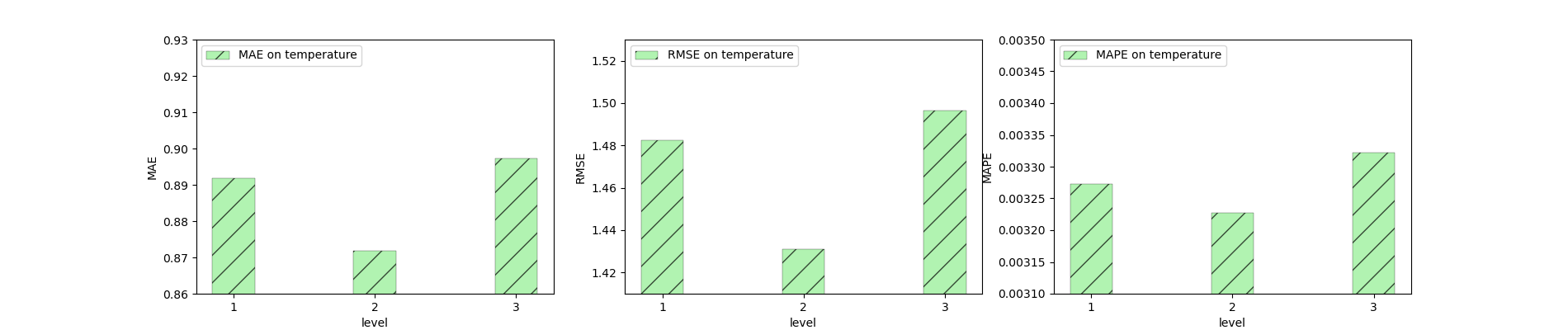}}\vspace{-0.3cm}
\end{figure}


\end{document}